\documentclass[10pt,twocolumn,letterpaper]{article}

\usepackage[pagenumbers]{cvpr}   

\usepackage{xcolor}

\newcommand{\vx}{\mathbf{x}}
\newcommand{\vy}{\mathbf{y}}
\newcommand{\timestep}{t}
\newcommand{\timesteps}{T}
\newcommand{\denoiser}{D}
\newcommand{\denoisedpatch}{\hat{\mathbf{x}}}
\newcommand{\blendedpatch}{\tilde{\mathbf{x}}}
\newcommand{\noisypatch}{\mathbf{x}}
\newcommand{\patch}{\mathbf{y}}
\newcommand{\corruptpatch}{{\mathbf{x}}}
\newcommand{\numpixels}{M}
\newcommand{\numcorruptpatches}{N}
\newcommand{\numpatches}{N}
\newcommand{\imageout}{\mathbf{x}}
\newcommand{\dataset}{\mathcal{Y}}

\newcommand{\numscales}{S}
\newcommand{\corruptdataset}{\mathcal{X}}
\newcommand{\noise}{\boldsymbol{\epsilon}}
\newcommand{\normal}{\mathcal{N}}
\newcommand{\getpatch}{\mathbf{P}}
\newcommand{\putpatch}{\mathbf{R}}

\newcommand{\mixingcoef}{\pi}

\definecolor{cvprblue}{rgb}{0.21,0.49,0.74}
\usepackage[pagebackref,breaklinks,colorlinks,allcolors=cvprblue]{hyperref}
\usepackage[accsupp]{axessibility}

\def\paperID{6149} 
\def\confName{CVPR}
\def\confYear{2026}

\title{Efficient and Training-Free Single-Image Diffusion Models}

\author{
  Haojun Qiu\textsuperscript{1} \quad
  Kiriakos N. Kutulakos\textsuperscript{1,2} \quad
  David B. Lindell\textsuperscript{1,2} \\
  \textsuperscript{1}Department of Computer Science, University of Toronto \quad
  \textsuperscript{2}Vector Institute \\
  \small{\url{https://haojunqiu.github.io/efficient-SID/}\vspace{-1em}}
}

\newcommand{\R}{\mathbb{R}}
\usepackage{comment}
\usepackage{xr}
\usepackage{wrapfig}
\usepackage{lipsum}
\usepackage{algorithm} 
\usepackage{algpseudocode} 
\usepackage{cuted} 

\usepackage{colortbl}
\usepackage{makecell}   
\usepackage{rotating}   
\usepackage{float}
\usepackage{multirow}
\usepackage{nameref}
\usepackage[acronym]{glossaries}

\usepackage{overpic} 
\usepackage{tocloft}

\makeatletter
\def\onecolumnnow{%
  \global\columnwidth=\textwidth
  \global\hsize=\columnwidth
  \global\linewidth=\columnwidth
  \global\@twocolumnfalse
  \global\@firstcolumnfalse
  \col@number\@ne
}

\makeatletter
\renewcommand{\l@section}    {\@dottedtocline{1}{0em}{2.0em}}
\renewcommand{\l@subsection} {\@dottedtocline{2}{1.5em}{2.8em}}
\renewcommand{\l@subsubsection}{\@dottedtocline{3}{3.0em}{4.8em}}
\makeatother

\begin{document}
\addtocontents{toc}{\protect\setcounter{tocdepth}{-1}} 

\maketitle

\renewcommand{\baselinestretch}{0.97}

\begin{abstract}
We consider the problem of generating images whose internal structure---defined by the distribution of patches across multiple scales---matches that of a single reference image. 
Recent approaches address this problem by training a diffusion model on a single image. 
But even in this setting, training is computationally expensive and requires hours of optimization.
Instead, we model the image using a dataset of its patches at different scales.
As this dataset is finite and the dimensionality of its patches is small, the score function for a noisy patch can be computed tractably using an optimal, closed-form denoiser, eliminating the need for neural network training.
We integrate this patch-based denoiser into an efficient, training-free image diffusion model, and we describe how our method connects to classical patch-based image restoration techniques.
Our approach achieves state-of-the-art generation quality and diversity compared to trained single-image diffusion models, and we demonstrate applications, including unconditional image generation, text-guided stylization, image symmetrization, and retargeting.
Further, we show that our approach is compatible with latent space diffusion, and we show multiple additional acceleration techniques to achieve megapixel single-image generation in one second, and gigapixel generation in minutes. 
\vspace{-1.5em}
\end{abstract}
    
\section{Introduction}
\label{sec:intro}

A single image contains a dataset of thousands to millions of patches---local neighborhoods or groups of pixels---occurring across different positions and scales.
The distribution of image patches conveys information about the \textit{internal structure} of an image~\cite{zontak_internal_2011}; for example, most images have patches that are self-similar within a scale, correlated in appearance across scales, and similar in their spatial frequency content. 
Analyzing and modeling the internal structure of images has led to significant advances in applications such as unconditional image generation~\cite{shaham_singan_2019,granot_drop_2021}, image manipulation~\cite{kulikov_sinddm_2022,nikankin_sinfusion_2023}, and image restoration~\cite{dabov2007image,mairal2009non,zoran2011learning,shocher_zero-shot_2018}.  

Although non-parametric sampling methods for patch-based image synthesis and restoration have a long history in computer vision~\protect\cite{efros1999texture,efros2001image,wei2001texture,jojic2003epitomic,buades2005non,dabov2007image,barnes2009patchmatch,glasner2009super}, recent work has focused instead on using generative models to learn the distribution of patches from a single image.
Techniques based on generative adversarial networks (GANs) generate images whose distribution of patches matches that of a source image based on the output of a discriminator~\cite{goodfellow2014generativeadversarialnetworks,shaham_singan_2019,hinz2021improved}.
Other techniques train a diffusion model to denoise a single image at multiple scales with varying amounts of Gaussian noise~\cite{kulikov_sinddm_2022,nikankin_sinfusion_2023}. 
After training, new images with a similar internal structure can be sampled by applying a coarse-to-fine denoising procedure.
However, training single-image generative models is computationally expensive, requiring several hours of optimization even though the training data comprises only a single image. 
Further, such generative models can be difficult to optimize---especially GANs, which are susceptible to local minima and mode collapse~\cite{miyato2018spectral,razavi2019generating,brock2019large,dhariwal2021diffusion}.

A key advantage of classical patch-based modeling techniques  is that they require no training, and thus are far more computationally lightweight compared to GANs and diffusion models. 
    Moreover, recent work in this direction has shown competitive results in terms of generated image quality~\cite{granot_drop_2021} despite relying
on nearest-neighbor patch matching~\cite{barnes2009patchmatch} instead of internal learning. 
Similar to denoising diffusion models, these methods generate images through a coarse-to-fine processing procedure that begins from a noisy, coarse-resolution input. 
But, instead of denoising, images are generated by
iteratively refining patches with nearest-neighbor matching and progressive upscaling.
While this approach is efficient and effective, it is less flexible than score-based denoising diffusion models~\protect\cite{ho_denoising_2020,song_ddim_2022}, which explicitly model the prior probability of image patches. 
For example, diffusion models are easily combined with vision--language models for text-based editing~\cite{kulikov_sinddm_2022,radford2021learning}, and they can incorporate symmetry constraints or local edits during the generation process~\cite{madar2024tiled}.

\begin{figure*}[ht]
    \centering
    \includegraphics[width=\textwidth]{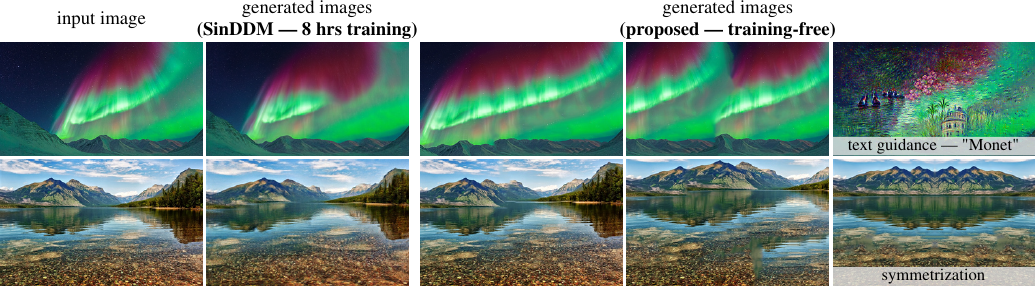} 
    \vspace{-2em}
    \captionof{figure}{We introduce an efficient, training-free diffusion model that generates images based on the internal structure of a single input image. Our approach uses a closed-form solution for the optimal denoiser derived for noisy patches of the input image. As such, no training is required, and generated images achieve similar or better quality (based on single-image Fréchet Inception Distance~\cite{shaham_singan_2019}) relative to state-of-the-art single-image diffusion models such as SinDDM~\cite{kulikov_sinddm_2022}, which require hours of training time. Our approach is also compatible with text-based guidance from pre-trained vision--language models~\cite{radford2021learning} and enables controllable generation, e.g., of symmetric images.}
\label{fig:teaser}
\vspace{-1.5em}
\end{figure*}

Here, we introduce a method for modeling the internal structure of images using a closed-form denoising diffusion procedure that is entirely training-free, similar to conventional patch-based methods. 
Hence, our approach avoids the computational cost of internal learning while inheriting the advantages of diffusion models in terms of explicit probabilistic modeling (see Figure~\ref{fig:teaser}). 

Our approach is based on the following key observation: since the set of patches in a single image is finite, 
the score function corresponding to the distribution of patches at all positions and scales
can be computed in closed form~\cite{karras_elucidating_2022,scarvelis_closed-form_2023,niedoba2025mechanisticexplanationdiffusionmodel,buchanan2025edgememorizationdiffusionmodels,lukoianov2025localityimagediffusionmodels,song2025selectiveunderfittingdiffusionmodels}, without training a neural network.
That is, evaluating the score function for a noisy patch corresponds to applying a denoiser that is optimal for the ensemble of patches in an image.
At the image level, the optimal denoiser takes a form similar to a non-local-means denoiser, 
thereby drawing a connection between recent denoising diffusion models~\cite{ho_denoising_2020,song_ddim_2022} and classical patch-based methods~\cite{buades2005non,zoran2011learning}. 
To generate coherent images across patch boundaries, we integrate this closed-form denoiser into a novel reverse diffusion process that operates in a coarse-to-fine fashion. 
Finally, we demonstrate our efficient, closed-form single-image denoiser for applications such as unconditional image generation, retargeting, text-based stylization and editing, and image symmetrization~\cite{mitra2007symmetrization,madar2024tiled}.

Although the connection between the score function and denoising has long been known~\cite{hyvarinen2005estimation,vincent2011connection}, we show that the analytical denoising solution is especially well-suited to modeling internal image structure. 
Moreover, we find that patch-based diffusion is amenable to multiple acceleration techniques: 
(1) fused attention kernels originally developed for transformers~\cite{vaswani2017attention,dao2022flashattention}, 
(2) latent space diffusion~\cite{rombach2022high,esser2024scaling},
and (3) approximate nearest neighbors for rapidly identifying similar patches~\cite{indyk1998approximate}.
Together, these techniques enable us to achieve megapixel generation in one second and gigapixel generation in minutes.
Overall, our closed-form denoising solution achieves state-of-the-art capabilities in single-image modeling 
without the 
hours-long training 
of other diffusion-based methods~\cite{kulikov_sinddm_2022,nikankin_sinfusion_2023}. 

\vspace{-0.5em}
\section{Related Work}
\label{sec:related work}

\paragraph{Classical patch-based models.} 

Modeling image structure using patches has long been attractive because it leads to tractable methods for image analysis, inference, and likelihood estimation. 
For example, analyzing the distribution of patches within natural images reveals that patches typically recur many times~\cite{zontak_internal_2011}. 
Patch self-similarity is the key principle of non-parametric single-image techniques for texture synthesis~\cite{efros1999texture,efros2001image,jacobs2001image}, stylization~\cite{barnes2009patchmatch,jacobs2001image}, restoration~\cite{elad2006image,dabov2007image}, and state-of-the-art non-local denoising methods~\cite{buades2005non,dabov2007image}.
Parametric methods seek to explicitly model the prior probability of patches using models such as Gaussian mixtures~\cite{zoran2011learning,papyan2015multi}.  
As we will show, our approach connects classical non-parametric and parametric techniques for patch-based modeling and grounds them in the modern framework of diffusion-based inference. 

\vspace{-1.0em}
\paragraph{Single-image GANs.} 
More recently, GANs have been used to learn the distribution of patches from a single image~\cite{shaham_singan_2019,shocher_ingan_2019,hinz2021improved}.
These methods train a generator to create images whose patch statistics match those of the input image across multiple scales. 
However, GANs do not support guidance (e.g., from text prompts) without re-training. 

\vspace{-1.0em}
\paragraph{Single-image diffusion models.}
Diffusion models are an attractive alternative to GANs; they are easier to train, achieve higher-quality generation~\cite{dhariwal2021diffusion}, and avoid challenges related to sample diversity~\cite{miyato2018spectral,razavi2019generating}.
In the context of single-image modeling, recent methods based on diffusion models show compelling results in image generation, manipulation, and text-driven stylization~\cite{kulikov_sinddm_2022,wang2022sindiffusion,nikankin_sinfusion_2023} 
but are expensive to train. Moreover, they model patches implicitly
by restricting the receptive field of the network~\cite{wang2022sindiffusion,nikankin_sinfusion_2023} or by generating images sequentially across scales~\cite{kulikov_sinddm_2022}.
Similar to our method, concurrent work investigates patch-based models in the context of closed-form diffusion, but focuses on texture synthesis rather than image generation~\cite{chatillon2025nifty}.
Our method operates explicitly on patches using an optimal closed-form denoiser, achieving similar or better image generation quality relative to prior training-based diffusion models---without any training.
See Sec.~\ref*{sec:supp-connections-to-other-sin-gen} for a detailed discussion of how our approach compares to previous work.

\section{Method}
\label{sec:method}

We now describe our method for single-image generative modeling by closed-form denoising diffusion. We begin with an overview
of diffusion sampling and the closed-form denoiser (\cref{sec:denoising-diffusion}) and  
connect our approach to classical patch-based methods (\cref{sec:patch-methods}).
We then describe how to integrate the closed-form denoiser into a multi-scale approach for image sampling
(\cref{sec:single-scale,sec:multi-scale}).

\begin{figure*}[h!] 
    \centering
    \begin{overpic}[width=\textwidth]{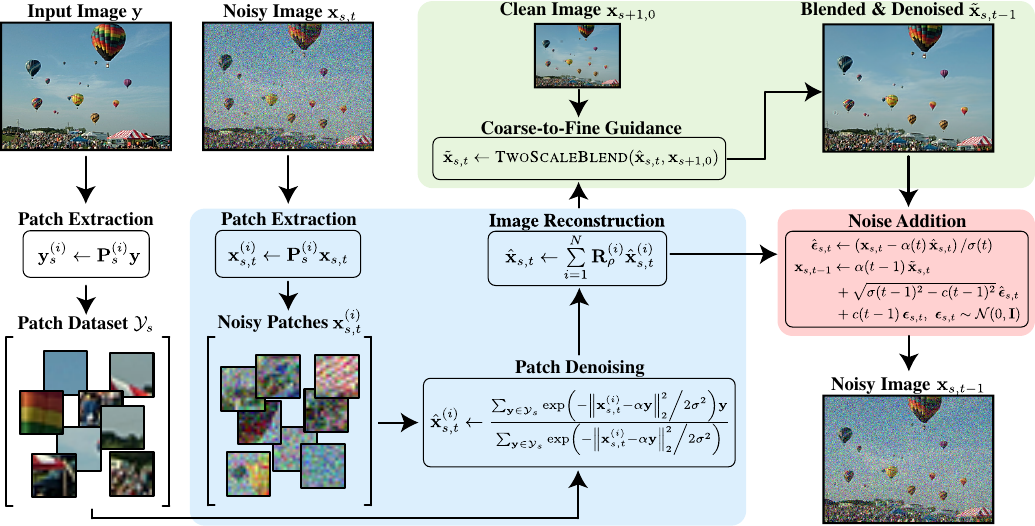}
    \end{overpic}\\
    {\small\textbf{Single-Scale Image Sampling}}
    \\
    \begin{overpic}[width=\textwidth]{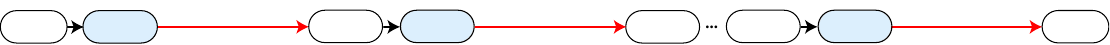}
    \setlength{\fboxsep}{0pt}
    \put(0.5,1){\parbox[c][10pt][c]{25pt}{\centering\scalebox{0.8}{$\noisypatch_{\timesteps}$}}}
    \put(8.24,1){\parbox[c][10pt][c]{25pt}{\centering\scalebox{0.8}{$\denoisedpatch_{\timesteps}$}}}
    \put(18.24,2.3){\parbox[c][10pt][c]{25pt}{\centering\scalebox{0.6}{\color{red}{$\timestep\!=\!\timesteps\!-\!1$}}}}
    \put(28.58,1){\parbox[c][10pt][c]{25pt}{\centering\scalebox{0.8}{$\noisypatch_{\timesteps-1}$}}}
    \put(36.82,1){\parbox[c][10pt][c]{25pt}{\centering\scalebox{0.8}{$\denoisedpatch_{\timesteps-1}$}}}
    \put(46.82,2.3){\parbox[c][10pt][c]{25pt}{\centering\scalebox{0.6}{\color{red}{$\timestep\!=\!\timesteps\!-\!2$}}}}
    \put(57.16,1){\parbox[c][10pt][c]{25pt}{\centering\scalebox{0.8}{$\noisypatch_{\timesteps-2}$}}}
    \put(66.26,1){\parbox[c][10pt][c]{25pt}{\centering\scalebox{0.8}{$\noisypatch_{1}$}}}
    \put(74.5,1){\parbox[c][10pt][c]{25pt}{\centering\scalebox{0.8}{$\denoisedpatch_{1}$}}}
    \put(84.5,2.3){\parbox[c][10pt][c]{25pt}{\centering\scalebox{0.6}{\color{red}{$\timestep\!=\!0$}}}}
    \put(94.4,1){\parbox[c][10pt][c]{25pt}{\centering\scalebox{0.8}{$\noisypatch_{0}$}}}
    \end{overpic}\\
        {\small\textbf{Coarse-to-Fine Image Sampling}}
    \\[2pt]
    \begin{overpic}[width=\textwidth]{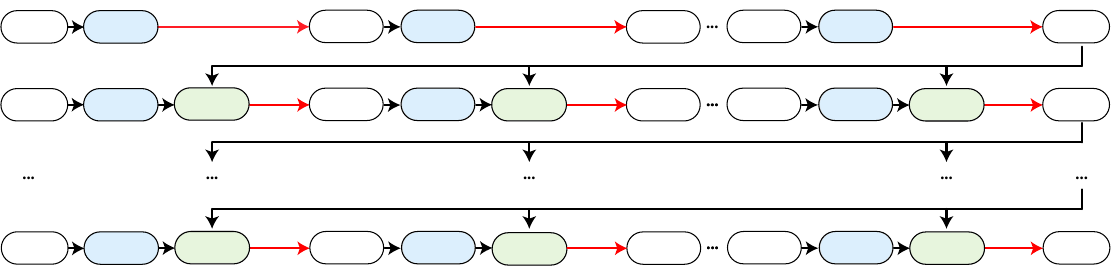}
    \setlength{\fboxsep}{0pt}
    \put(0.5,21.5){\parbox[c][10pt][c]{25pt}{\centering\scalebox{0.7}{$\noisypatch_{\numscales,\timesteps}$}}}
    \put(8.24,21.5){\parbox[c][10pt][c]{25pt}{\centering\scalebox{0.7}{$\denoisedpatch_{\numscales,\timesteps}$}}}
    \put(18.24,22.8){\parbox[c][10pt][c]{25pt}{\centering\scalebox{0.6}{\color{red}{$\timestep\!=\!\timesteps\!-\!1$}}}}
    \put(28.58,21.5){\parbox[c][10pt][c]{25pt}{\centering\scalebox{0.7}{$\noisypatch_{\numscales,\timesteps-1}$}}}
    \put(36.82,21.5){\parbox[c][10pt][c]{25pt}{\centering\scalebox{0.7}{$\denoisedpatch_{\numscales,\timesteps-1}$}}}
    \put(46.82,22.8){\parbox[c][10pt][c]{25pt}{\centering\scalebox{0.6}{\color{red}{$\timestep\!=\!\timesteps\!-\!2$}}}}
    \put(57.16,21.5){\parbox[c][10pt][c]{25pt}{\centering\scalebox{0.7}{$\noisypatch_{\numscales,\timesteps-2}$}}}
    \put(66.26,21.5){\parbox[c][10pt][c]{25pt}{\centering\scalebox{0.7}{$\noisypatch_{\numscales,1}$}}}
    \put(74.5,21.5){\parbox[c][10pt][c]{25pt}{\centering\scalebox{0.7}{$\denoisedpatch_{\numscales,1}$}}}
    \put(84.5,22.8){\parbox[c][10pt][c]{25pt}{\centering\scalebox{0.6}{\color{red}{$\timestep\!=\!0$}}}}
    \put(94.4,21.5){\parbox[c][10pt][c]{25pt}{\centering\scalebox{0.7}{$\noisypatch_{\numscales,0}$}}}
    \put(0.5,14.5){\parbox[c][10pt][c]{25pt}{\centering\scalebox{0.7}{$\noisypatch_{\numscales-1,\timesteps}$}}}
    \put(8.24,14.5){\parbox[c][10pt][c]{25pt}{\centering\scalebox{0.7}{$\denoisedpatch_{\numscales-1,\timesteps}$}}}
    \put(16.48,14.5){\parbox[c][10pt][c]{25pt}{\centering\scalebox{0.7}{$\blendedpatch_{\numscales-1,\timesteps}$}}}
    \put(22.3,15.8){\parbox[c][10pt][c]{25pt}{\centering\scalebox{0.6}{\color{red}{$\timestep\!=\!\timesteps\!-\!1$}}}}
    \put(28.58,14.5){\parbox[c][10pt][c]{25pt}{\centering\scalebox{0.7}{$\noisypatch_{\numscales-1,\timesteps-1}$}}}
    \put(36.82,14.5){\parbox[c][10pt][c]{25pt}{\centering\scalebox{0.7}{$\denoisedpatch_{\numscales-1,\timesteps-1}$}}}
    \put(45.06,14.5){\parbox[c][10pt][c]{25pt}{\centering\scalebox{0.7}{$\blendedpatch_{\numscales-1,\timesteps-1}$}}}
    \put(51.06,15.8){\parbox[c][10pt][c]{25pt}{\centering\scalebox{0.6}{\color{red}{$\timestep\!=\!\timesteps\!-\!2$}}}}
    \put(57.16,14.5){\parbox[c][10pt][c]{25pt}{\centering\scalebox{0.7}{$\noisypatch_{\numscales-1,\timesteps-2}$}}}
    \put(66.26,14.5){\parbox[c][10pt][c]{25pt}{\centering\scalebox{0.7}{$\noisypatch_{\numscales-1,1}$}}}
    \put(74.5,14.5){\parbox[c][10pt][c]{25pt}{\centering\scalebox{0.7}{$\denoisedpatch_{\numscales-1,1}$}}}
    \put(82.74,14.5){\parbox[c][10pt][c]{25pt}{\centering\scalebox{0.7}{$\blendedpatch_{\numscales-1,1}$}}}
    \put(88.3,15.8){\parbox[c][10pt][c]{25pt}{\centering\scalebox{0.6}{\color{red}{$\timestep\!=\!0$}}}}
    \put(94.4,14.5){\parbox[c][10pt][c]{25pt}{\centering\scalebox{0.7}{$\noisypatch_{\numscales-1,0}$}}}
    \put(0.5,1.6){\parbox[c][10pt][c]{25pt}{\centering\scalebox{0.7}{$\noisypatch_{0,\timesteps}$}}}
    \put(8.24,1.6){\parbox[c][10pt][c]{25pt}{\centering\scalebox{0.7}{$\denoisedpatch_{0,\timesteps}$}}}
    \put(16.48,1.6){\parbox[c][10pt][c]{25pt}{\centering\scalebox{0.7}{$\blendedpatch_{0,\timesteps}$}}}
    \put(22.3,2.9){\parbox[c][10pt][c]{25pt}{\centering\scalebox{0.6}{\color{red}{$\timestep\!=\!\timesteps\!-\!1$}}}}
    \put(28.58,1.6){\parbox[c][10pt][c]{25pt}{\centering\scalebox{0.7}{$\noisypatch_{0,\timesteps-1}$}}}
    \put(36.82,1.6){\parbox[c][10pt][c]{25pt}{\centering\scalebox{0.7}{$\denoisedpatch_{0,\timesteps-1}$}}}
    \put(45.06,1.6){\parbox[c][10pt][c]{25pt}{\centering\scalebox{0.7}{$\blendedpatch_{0,\timesteps-1}$}}}
    \put(51.06,2.9){\parbox[c][10pt][c]{25pt}{\centering\scalebox{0.6}{\color{red}{$\timestep\!=\!\timesteps\!-\!2$}}}}
    \put(57.16,1.6){\parbox[c][10pt][c]{25pt}{\centering\scalebox{0.7}{$\noisypatch_{0,\timesteps-2}$}}}
    \put(66.26,1.6){\parbox[c][10pt][c]{25pt}{\centering\scalebox{0.7}{$\noisypatch_{0,1}$}}}
    \put(74.5,1.6){\parbox[c][10pt][c]{25pt}{\centering\scalebox{0.7}{$\denoisedpatch_{0,1}$}}}
    \put(82.74,1.6){\parbox[c][10pt][c]{25pt}{\centering\scalebox{0.7}{$\blendedpatch_{0,1}$}}}
    \put(88.3,2.9){\parbox[c][10pt][c]{25pt}{\centering\scalebox{0.6}{\color{red}{$\timestep\!=\!0$}}}}
    \put(94.4,1.6){\parbox[c][10pt][c]{25pt}{\centering\scalebox{0.7}{$\noisypatch_{0,0}$}}}
    \end{overpic}
    \vspace*{-2em}
    \caption{Method overview. Our approach takes a single image as input, extracts patches, and uses the patches to generate new images using denoising diffusion. \textbf{(top)} We illustrate a single step of the reverse diffusion process: \textit{(blue)} patches from the noisy image are denoised and used to reconstruct an image; \textit{(green)} the denoised image is blended with the output of the reverse diffusion process at a coarser scale ($\noisypatch_{s+1, \timestep}$); \textit{(red)} the noisy image at the previous diffusion timestep ($\noisypatch_{s, \timestep-1}$) is sampled. \textbf{(bottom)} All steps of reverse diffusion process for single-scale image sampling (omits coarse-to-fine guidance (\textit{green})) and coarse-to-fine image sampling are shown.} 
        \label{fig:method} 
    \vspace*{-15pt}
\end{figure*}
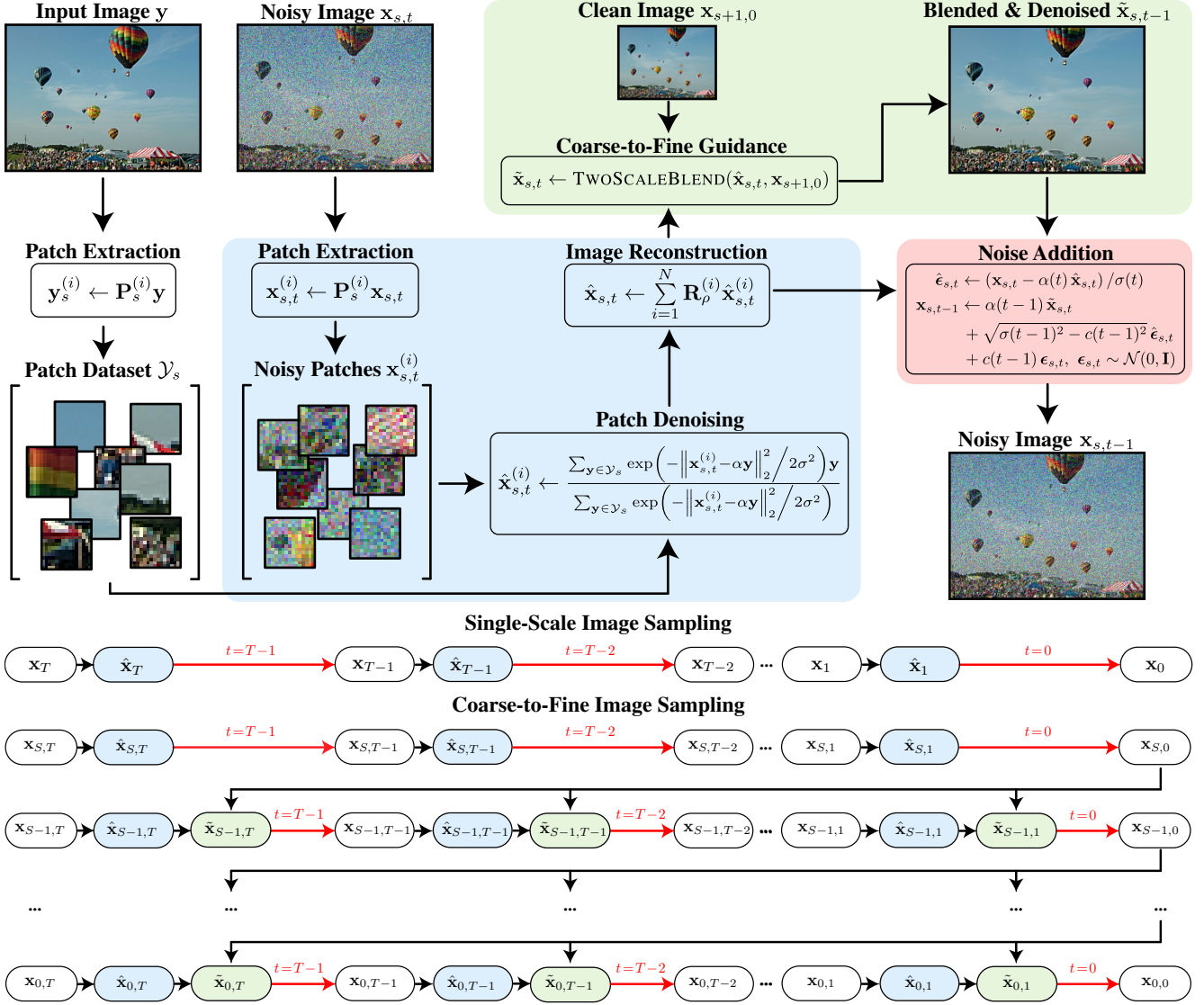

\subsection{Closed-Form Denoising Diffusion}
\label{sec:denoising-diffusion}

\paragraph{Preliminaries.}
Diffusion models~\cite{sohl-dickstein_deep_2015,ho_denoising_2020,song_generative_2020,song_ddim_2022,song_score-based_2021} act on a forward diffusion process that adds an increasing amount of noise in steps $\timestep \in [0, \ldots, \timesteps]$ to a clean signal $\patch\in\mathbb{R}^{\numpixels}$ sampled from a dataset $\dataset$.
The noisy signal $\noisypatch_\timestep$ at step $\timestep$ is given as $\noisypatch_\timestep = \alpha(t) \patch + \sigma(\timestep) \noise$, where $\noise \sim \normal(\mathbf{0}, \mathbf{I})$ is standard normal distributed, and the noise-scheduling parameters $\alpha(t)$, $\sigma(t)$ are smooth functions chosen so that $\alpha(0) = \sigma(\timesteps) = 1$ and $\alpha(\timesteps) = \sigma(0) = 0$ (i.e., $\timestep=0$ corresponds to the clean signal).
The diffusion model learns the structure of signals in $\dataset$ by running the
diffusion process in reverse. That is, the model consists of a denoiser $\denoiser$ that is trained to minimize
\vspace{-0.5em}
\begin{equation}
    \mathbb{E}_{
        \patch \sim \dataset, 
        \timestep \sim \mathcal{U}[0,\timesteps], 
        \noise \sim \normal(\mathbf{0}, \mathbf{I})}
    \left[  
        w(\timestep)
        \|
        \denoiser(
            \noisypatch_\timestep, 
            \timestep
        )
        - \patch
        \|_{2}^{2}
    \right],
    \label{eq:diffusion objective}
\vspace{-0.5em}
\end{equation}
where $w(\timestep)$ is a weight that depends on $\timestep$~\cite{song_ddim_2022}. 
Hence, the model learns to denoise all signals in the dataset over all diffusion timesteps.

\vspace{-1em}
\paragraph{Closed-form denoising.}
Instead of training the diffusion model, our approach uses an optimal, closed-form denoiser 
inspired by previous work~\cite{karras_elucidating_2022,scarvelis_closed-form_2023,niedoba2025mechanisticexplanationdiffusionmodel,kamb2025an,bertrand2025on,buchanan2025edgememorizationdiffusionmodels,lukoianov2025localityimagediffusionmodels,song2025selectiveunderfittingdiffusionmodels}.
We use this denoiser to iteratively reverse the diffusion process to recover a clean signal from noise.
The closed-form denoiser is given as 
\vspace{-1em}
\begin{equation}
    \denoiser(\noisypatch_\timestep, \dataset, \timestep) = \frac{\sum\limits_{\patch\in\dataset}p_\normal(\noisypatch_\timestep; \alpha\patch, \sigma^2\mathbf{I})\,\patch}{\sum\limits_{\patch\in\dataset}p_\normal(\noisypatch_\timestep; \alpha\patch, \sigma^2\mathbf{I})}, 
    \label{eq:closed-form}
    \vspace{-0.5em}
\end{equation}
which computes the denoised signal as a weighted function of all clean signals in the dataset $\dataset = \{\patch^{(1)}, \ldots \patch^{(Y)} \}$ (please see~\cref{sec:supp-closed-form} for the derivation).
We use the denoiser to run the reverse diffusion process with the following iterative updates:
\begin{align}
    \hat{\mathbf{x}}_t 
    &\gets \denoiser(\noisypatch_\timestep, \dataset, \timestep) &\text{(denoised signal)},
    \label{eq:DDIM-step-1}
    \\
    \hat{\boldsymbol\epsilon}_t &\gets (\noisypatch_\timestep - \alpha(\timestep) \denoisedpatch_\timestep) / \sigma(\timestep) &\text{(estimated noise)}, 
    \label{eq:DDIM-step-2}
    \\
    {\noisypatch_{\timestep-1}} &\gets \alpha(\timestep-1) \denoisedpatch_{\timestep} &\text{(noisy signal)}, 
    \label{eq:DDIM-step-3} 
    \\ &\,\, + \sqrt{\sigma(\timestep-1)^2 - c(\timestep-1)^2} \hat{\boldsymbol \epsilon}_{t} 
    + c(\timestep-1) \boldsymbol{\epsilon}_t \hspace{-11em}& \notag 
    &     \\
    t &\gets t - 1 & \text{(update timestep)}.
    \label{eq:DDIM-step-4}
\end{align}
where $c(t-1) \in [0, \sigma(t-1)]$ modifies the amount of sampling stochasticity by adding random noise $\boldsymbol{\epsilon}_t \sim \mathcal{N}(\mathbf{0}, \mathbf{I})$, and we use $\eta(t) = c(t) / \sigma(t) \in [0, 1]$ to control the stochasticity. 
Equations~\ref{eq:DDIM-step-1}--\ref{eq:DDIM-step-4} reduce to a deterministic Denoising Diffusion Implicit Model (DDIM) \cite{song_ddim_2022} when $\eta(t)=0$.
Iterating these steps yields the generated signal $\noisypatch_{0}$.

\subsection{Connection to Patch-Based Image Restoration}
\label{sec:patch-methods}

Many classical
image restoration methods that rely on non-parametric sampling~\cite{buades2005non} can be viewed as employing the exact same closed-form denoiser at the patch level, followed by an image reconstruction step that reassembles a full image from a set of denoised patches.
Below, we describe connections of our framework to these classical techniques.

\vspace{-1em}
\paragraph{Non-local means denoising~\cite{buades2005non}.}
Given an input noisy image $\corruptpatch$, non-local means computes a denoised image $\denoisedpatch$ from a dataset of all overlapping noisy image patches, \mbox{$\corruptdataset=\{\getpatch^{(1)}, \ldots, \getpatch^{(\numcorruptpatches)} \corruptpatch  \} \overset{\mathrm{def}}{=}\{\corruptpatch^{(1)}, \ldots,  \corruptpatch^{(\numcorruptpatches)}\}$}, where $\getpatch^{(i)}$ is a matrix that extracts a patch $\corruptpatch^{(i)}$ from the image.
Each patch is denoised as  
    \vspace{-1em}
\begin{equation}
    \underbrace{\hphantom{x}\vphantom{\{}\denoisedpatch^{(i)}\hphantom{x}}_{\text{denoised patch}} \gets~~ \denoiser(\hspace*{-10pt}\underbrace{\hphantom{x}\vphantom{\{}\corruptpatch^{(i)}\hphantom{x}}_{\text{patch to be denoised}}\hspace*{-10pt}, ~~\quad \underbrace{\hphantom{.}\corruptdataset - \{\corruptpatch^{(i)}\}\hphantom{.}}_{\text{all other noisy patches}}, ~~\timestep ).
    \label{eq:nlm}
    \vspace{-1em}
\end{equation}
The output image is then assembled as $\denoisedpatch \gets \sum_{i=1}^{\numcorruptpatches} \putpatch^{(i)} \denoisedpatch^{(i)}$, where $\putpatch^{(i)}$ is a matrix that copies the center pixel of the patch back to its corresponding location in the image. 
Note that the formulation here is identical to Equation~\ref{eq:closed-form}, except that the dataset $\corruptdataset$ consists of noisy patches.

\vspace{-1em}
\paragraph{Image restoration with GMM patch priors~\cite{zoran2011learning}.}

One way to model patch priors for image restoration is to fit a Gaussian mixture model (GMM) to the patch dataset~\cite{zoran2011learning}. 
This results in a prior patch probability of \mbox{$p(\corruptpatch^{(i)}) = \sum\limits_{k=1}^{K} \mixingcoef_k p_\normal(\imageout^{(i)}; \boldsymbol{\mu}_k, \boldsymbol{\Sigma}_k)$}, where $\mixingcoef_k$ are the mixing coefficients for the $K$ components, and $\boldsymbol{\mu}_k$ and $\boldsymbol{\Sigma}_k$ are the mean and covariance matrix, respectively.
However, this prior does not yield a closed-form solution for the maximum a posteriori estimate of the clean patch
given a noisy patch $\corruptpatch^{(i)}$~\cite{zoran2011learning}.
Further, fitting the GMM prior to the patch dataset requires an expensive optimization using expectation maximization.

The closed-form denoiser of Equation~\ref{eq:closed-form} can be thought of as a restoration technique that replaces the above patch prior with a trivial GMM that centers a mixture component at each patch in the dataset: \mbox{$p(\corruptpatch^{(i)}) =  \frac{1}{\numpatches}\sum_{j=1}^{\numpatches} p_\mathcal{N}(\corruptpatch^{(i)}; \patch^{(j)}, \sigma^2\mathbf{I})$.} As $\sigma \rightarrow 0$, Equation~\ref{eq:closed-form} converges to the exact prior probability $p(\corruptpatch^{(i)})$, and the closed-form denoiser gives the minimum mean squared error estimate of a clean patch given the noisy patch $\corruptpatch^{(i)}$ (see~\cref{sec:supp-closed-form}).
Hence, our formulation relies on a GMM prior---just like classical patch-based methods---but one that estimates clean patches efficiently in closed form.

\begin{algorithm}[h!]\small
\caption{Single-Scale Image Sampling}
\label{alg:single-scale-sampling}
\begin{algorithmic}[1]
\Procedure{SampleImage}{$\patch$}
\State $ \dataset = \{ \getpatch^{(1)}\patch,\ldots, \getpatch^{(\numpatches)}\patch \} $ \Comment{extract clean patches}
\State $\noisypatch_\timesteps \sim \mathcal{N}(\mathbf{0}, \mathbf{I})$
\For{$\timestep = [\timesteps,  \ldots, 1]$} \Comment{reverse diffusion steps} 
        \State $\denoisedpatch_{t} 
    \leftarrow  \Call{ImgDenoise}{\noisypatch_\timestep, \dataset, \timestep}$
        \State $ \hat{\boldsymbol\epsilon}_\timestep \leftarrow (\noisypatch_\timestep - \alpha(\timestep) \denoisedpatch_\timestep) / \sigma(\timestep)$
        \State $\begin{aligned}
                \noisypatch_{\timestep-1} 
                \leftarrow {} 
                \alpha(\timestep-1) \denoisedpatch_{\timestep} 
                & + \sqrt{\sigma(\timestep-1)^2 - c(\timestep-1)^2} \hat{\boldsymbol \epsilon}_{t} \\
                & + c(\timestep-1) \boldsymbol{\epsilon}_t, \quad \boldsymbol{\epsilon}_t \sim \mathcal{N}(\mathbf{0}, \mathbf{I})
                \end{aligned}
                $
    \EndFor
    \State \Return $\noisypatch_{t=0}$
\EndProcedure

\Procedure{ImgDenoise}{$\noisypatch_{t}, \dataset, \timestep$}
\State $ \{\noisypatch_\timestep^{(i)} \}_{i=1}^{\numcorruptpatches} \leftarrow \{ \getpatch^{(1)}\noisypatch_{t},\ldots, \getpatch^{(\numcorruptpatches)}\noisypatch_{t} \} $
\Comment{extract patches}
\State $\{ \denoisedpatch_{\timestep}^{(i)} \}_{i=1}^{\numcorruptpatches} 
        \leftarrow  \{ \Call{PatchDenoise}{\noisypatch_{\timestep}^{(i)}, \dataset, \timestep} \}_{i=1}^{\numcorruptpatches}$ \hspace*{-15pt}
        \Comment{denoise}
\State $\denoisedpatch_\timestep \leftarrow \sum_{i=1}^{\numcorruptpatches} \putpatch_\rho^{(i)}\denoisedpatch^{(i)}_{\timestep}$
\Comment{reconstruct image}
\State \Return $\denoisedpatch_\timestep$
\EndProcedure

\Procedure{PatchDenoise}{$\noisypatch_{t}, \dataset, \timestep$}
\State $\denoisedpatch_\timestep \leftarrow \frac{\sum_{\patch \in \dataset} \exp \left( -\|\noisypatch_{t} - \alpha\patch \|_{2}^{2} \middle/ 2 \sigma^2 \right) \,\patch}{\sum_{\patch \in \dataset} \exp \left( -\|\noisypatch_{t} - \alpha\patch \|_{2}^{2} \middle/ 2 \sigma^2 \right)}  $
    \State \Return $\denoisedpatch_{\timestep}$
\EndProcedure

\end{algorithmic}
\end{algorithm}

\subsection{Single-Scale Image Sampling}
\label{sec:single-scale}

We now
describe how to sample entire images by applying the closed-form denoiser at the patch level.
The procedure is illustrated in Figure~\ref{fig:method} and
summarized in Algorithm~\ref{alg:single-scale-sampling}. 

\begin{figure}[h]
    \vspace{-1em}
    \includegraphics[width=\columnwidth]{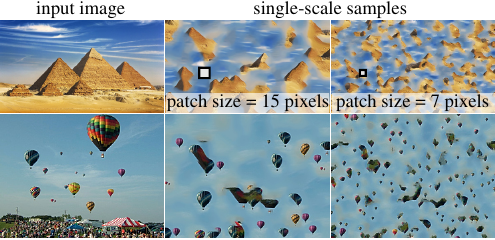}
    \vspace{-2em}
\caption{Illustration of single-scale sampling. This procedure (detailed in Algorithm~\ref{alg:single-scale-sampling}) captures image statistics at the scale of an individual patch (white squares),  but fails to capture the coarse structure of the image. We address this issue using coarse-to-fine image sampling (Algorithm~\ref{alg:coarse-to-fine}, Figure~\ref{fig:unconditional}).}
    \label{fig:single-scale}
\end{figure}

The procedure begins by extracting a dataset $\dataset$ of overlapping patches from the input image. These patches are treated as ``clean,'' i.e., noise-free. To sample an image,
 we start at the last timestep of the forward diffusion
  by sampling a noise image $\noisypatch_{t=\timesteps}\sim\normal(\mathbf{0}, \mathbf{I})$.
  We then
 iteratively denoise it
  through the reverse diffusion process as follows.
We extract noisy patches from the image as $\noisypatch^{(i)}_\timestep \gets \getpatch^{(i)}\noisypatch_\timestep$, and we apply Equations~\ref{eq:DDIM-step-1}--\ref{eq:DDIM-step-4} to recover denoised patches $\denoisedpatch_t^{(i)}$.  
Then, we reconstruct a full image as $\denoisedpatch_{\timestep} \gets \sum_{i=1}^{\numpatches} \putpatch^{(i)}_{\rho} \denoisedpatch^{(i)}_{\timestep}$, where $\putpatch^{(i)}_\rho$ is a matrix that copies the patch back to its original location in the image after weighting by a Gaussian with standard deviation $\rho$.
Here, $\rho$ controls how much of the patch outside of the center pixel is copied back to the image (i.e., $\rho=0$ corresponds to the $\putpatch^{(i)}$ matrix employed by non-local-means denoising).   
Finally, we add noise to sample a noisy image for timestep $t \gets t-1$ of the forward diffusion. 
This process iterates until we sample the output image $\noisypatch_{t=0}$. Figure~\ref{fig:method} (middle) illustrates this sequence of denoising and sampling steps.

We show examples of single-scale image sampling in Figure~\ref{fig:single-scale}.
Note that while image structures at the scale of a patch are preserved, the sampled images do not maintain the input image's global structure. 
To address this issue, we develop a coarse-to-fine image sampling procedure that preserves global image structure from coarse scales while maintaining high-frequency details from fine scales.

\begin{algorithm}[t]\small
\renewcommand{\algorithmicindent}{1em} 
\caption{Coarse-to-Fine Image Sampling}\label{alg:cap}
\label{alg:coarse-to-fine}
\begin{algorithmic}[1]
    \Procedure{SampleImageCoarseToFine}{$\patch$}
    \State $ \{\dataset_s\}_{s=0}^\numscales \leftarrow \{ \getpatch_s^{(1)}\patch,\ldots, \getpatch_s^{({\numpatches_s})}\patch \}_{s=0}^{\numscales} $ \Comment{extract patches}
    \State \{$\noisypatch_{s, \timesteps}\}_{s=0}^{\numscales} \sim \normal(\mathbf{0}, \mathbf{I})$ \Comment{initialize a noise pyramid}
    \For{$s = [\numscales, \dots, 0]$} \Comment{iterate over scales} 
    \For{$\timestep = [\timesteps , \dots, 1] $}  \Comment{diffusion timesteps}
        \State $\denoisedpatch_{s, \timestep} \leftarrow \Call{ImgDenoise}{\noisypatch_{s, \timestep},\dataset_s, \timestep}$ 
        \State $ \hat{\boldsymbol\epsilon}_{s, \timestep} \leftarrow (\noisypatch_{s, \timestep} - \alpha(\timestep) \denoisedpatch_{s, \timestep}) / \sigma(\timestep)$
        \If{$s < \numscales$}
        \State $\blendedpatch_{s, \timestep} \leftarrow \Call{TwoScaleBlend}{\denoisedpatch_{s, \timestep}, \noisypatch_{s+1, t=0}}$
        \EndIf
        \State {\scalebox{0.93}{
                $\begin{aligned}
                \noisypatch_{s,\timestep-1} 
                \leftarrow {} 
                \alpha(\timestep-1) \blendedpatch_{s,\timestep} 
                & + \sqrt{\sigma(\timestep-1)^2 - c(\timestep-1)^2} \hat{\boldsymbol \epsilon}_{s,t} \\
                & + c(\timestep-1) \boldsymbol{\epsilon}_{s,t}, \; \boldsymbol{\epsilon}_{s,t} \sim \mathcal{N}(\mathbf{0}, \mathbf{I})
                \end{aligned}$}}
        \EndFor
        
    \EndFor
    \State \Return $\noisypatch_{s=0, \timestep=0}$
    \EndProcedure

    \Procedure{TwoScaleBlend}{$\denoisedpatch_{s, \timestep}, \noisypatch_{s+1, \timestep=0}$}
        \State \Return $ \denoisedpatch_{s, \timestep} - \textsc{Blur}(\denoisedpatch_{s, \timestep}) + \textsc{Upsample}(\noisypatch_{s+1, \timestep=0}) $
    \EndProcedure
\end{algorithmic}
\end{algorithm}

\begin{figure*}[t]
    \begin{minipage}{\textwidth}
  \centering
  \setlength{\tabcolsep}{1pt}
  \newcommand{\sz}{0.121}  %

  \begin{tabular}{cccccccc}
    {\small input image} & 
    {\small SinGAN \cite{shaham_singan_2019}} & 
    {\small GPNN \cite{granot_drop_2021}} & 
    {\small GPDM~\cite{elnekave2022generating}} &
    {\small SinDDM \cite{kulikov_sinddm_2022}} & 
    \small{ SinFusion \cite{nikankin_sinfusion_2023}} & 
    \small{
        \hspace{-5pt}
        SinDiffusion~\cite{wang2022sindiffusion}
        \hspace{-5pt}} & 
    \small{proposed} \\ 
    \makecell{\includegraphics[width=\sz\linewidth]{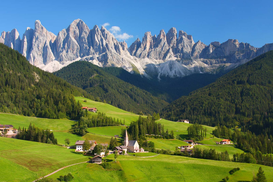}} &
    \makecell{\includegraphics[width=\sz\linewidth]{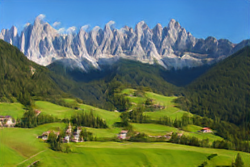}} & 
    \makecell{\includegraphics[width=\sz\linewidth]{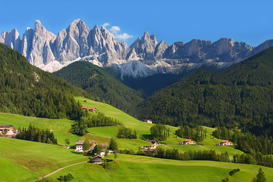}} &
    \makecell{\includegraphics[width=\sz\linewidth]{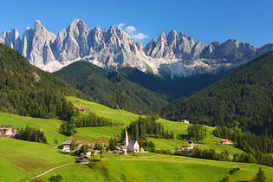}} &
    \makecell{\includegraphics[width=\sz\linewidth]{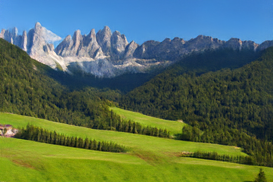}} &
    \makecell{\includegraphics[width=\sz\linewidth]{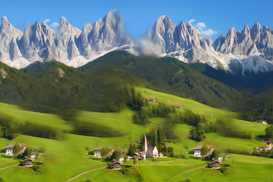}} & 
    \makecell{\includegraphics[width=\sz\linewidth]{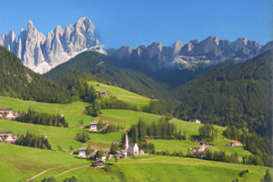}} & 
    \makecell{\includegraphics[width=\sz\linewidth]{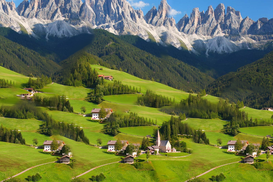}} \\[-0.2em]
    \makecell{\includegraphics[width=\sz\linewidth]{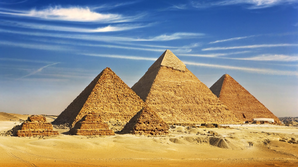}} &
    \makecell{\includegraphics[width=\sz\linewidth]{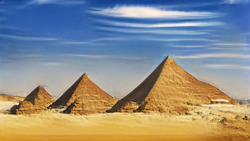}} & 
    \makecell{\includegraphics[width=\sz\linewidth]{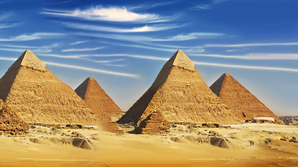}} &
    \makecell{\includegraphics[width=\sz\linewidth]{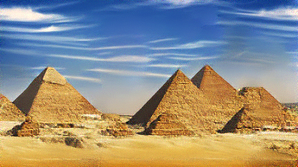}} &
    \makecell{\includegraphics[width=\sz\linewidth]{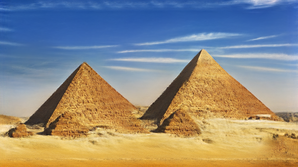}} &
    \makecell{\includegraphics[width=\sz\linewidth]{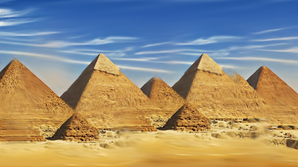}} & 
    \makecell{\includegraphics[width=\sz\linewidth]{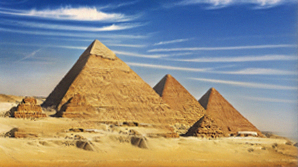}} & 
    \makecell{\includegraphics[width=\sz\linewidth]{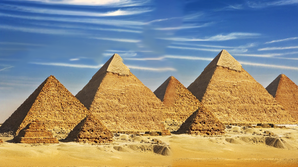}} \\ [-0.1em]
  \end{tabular}
  \vspace{-1em}
  \caption{Unconditional single-image generation results. Our training-free, coarse-to-fine image sampling procedure based on closed-form denoising diffusion (right) produces results of the same quality as other state-of-the-art methods that require hours of training time.}
  \label{fig:unconditional}
  \vspace{0.0em}
\end{minipage}

\begin{minipage}{\textwidth}
\captionsetup{justification=raggedright,singlelinecheck=false}
\centering

    \resizebox{\textwidth}{!}{
\begin{tabular}{ll|cccccccccc}
\toprule
{type} & {metric} 
 & {SinGAN~\cite{shaham_singan_2019}} 
 & {GPNN~\cite{granot_drop_2021}} 
 & {GPDM~\cite{elnekave2022generating}}
 & {SinDDM~\cite{kulikov_sinddm_2022}} 
 & {SinFusion~\cite{nikankin_sinfusion_2023}} 
 & {SinDiffusion~\cite{wang2022sindiffusion}}
 & {proposed ($T{=}10$,$\eta{=}0$)} 
 & {proposed ($T{=}40$, $\eta{=}1$)} 
 & {proposed ($T{=}10,\eta{=}0,k{=}5$)} \\\midrule
\textit{patch distribution} 
 & SIFID $\downarrow$
 & 0.13$\pm$0.08 
 & 0.06$\pm$0.11
 & \textbf{0.015$\pm$0.01}
 & 0.48$\pm$0.62 
 & 0.51$\pm$0.49 
 & 0.31$\pm$0.35
 & 0.29$\pm$0.39 
 & 0.21$\pm$0.29
 & 0.38$\pm$0.52
 \\
\midrule
\multirow{3}{*}{\textit{no reference IQA}}
 & NIQE $\downarrow$ 
 & 7.95$\pm$3.37
 & 9.78$\pm$5.59
 & 7.99$\pm$3.04
 & 7.69$\pm$3.60
 & 10.17$\pm$5.29
 & \textbf{6.96$\pm$2.88}
 & 8.08$\pm$3.23
 & 8.18$\pm$3.05
 & 8.10$\pm$3.50 
 \\
& NIMA $\uparrow$ 
 & 4.32$\pm$0.39
 & 4.69$\pm$0.48
 & 4.21$\pm$0.35
 & 4.30$\pm$0.46
 & \textbf{4.75$\pm$0.45}
 & 4.19$\pm$0.43
 & 4.53$\pm$0.45
 & 4.47$\pm$0.48
 & 4.52$\pm$0.43 
 \\
& MUSIQ $\uparrow$ 
 & 48.26$\pm$10.40
 & \textbf{56.60$\pm$11.04}
 & 49.72$\pm$11.41
 & 50.74$\pm$11.18
 & 51.38$\pm$12.40
 & 49.31$\pm$9.83
 & 55.41$\pm$11.05
 & 55.81$\pm$11.40
 & 55.13$\pm$11.32 
 \\
\midrule
\multirow{2}{*}{\textit{diversity}}
& Pixel Div.$\uparrow$ 
 & 0.09$\pm$0.03 
 & 0.08$\pm$0.02 
 & 0.10$\pm$0.04
 & 0.10$\pm$0.03 
 & 0.11$\pm$0.03 
 & 0.11$\pm$0.03
 & \textbf{0.15$\pm$0.04}
 & 0.13$\pm$0.03
 & 0.15$\pm$0.03 
\\
& LPIPS Div.$\uparrow$ 
 & 0.27$\pm$0.07 
 & 0.29$\pm$0.09
 & 0.31$\pm$0.14
 & 0.36$\pm$0.07 
 & 0.38$\pm$0.07 
 & 0.41$\pm$0.07
 & \textbf{0.49$\pm$0.07}
 & {0.39$\pm$0.06}
 & 0.50$\pm$0.08 
\\
\midrule
\multirow{2}{*}{\textit{training time} }
& TITAN RTX (hrs) $\downarrow$
 & 2.0
 & \textbf{0.0}
 & \textbf{0.0}
 & 10.0 
 & 3.2
 & 5.4
 & \textbf{0.0}
 & \textbf{0.0}
 & \textbf{0.0}
\\
& A6000 (hrs) $\downarrow$
 & \textit{not supported}
 & \textbf{0.0}
 & \textbf{0.0}
 & 8.0
 & 1.5
 & 4.2
 & \textbf{0.0}
 & \textbf{0.0} 
 & \textbf{0.0} \\
\midrule
\multirow{2}{*}{\textit{inference time} }
& TITAN RTX (s) $\downarrow$
& \textbf{0.04$\pm$0.00}
 & 2.62$\pm$0.01
 & 9.82$\pm$0.29
 & 1.60$\pm$0.06
 & 2.09$\pm$0.04
 & 14.25$\pm$0.23
 & 4.49$\pm$0.02
 & 18.53$\pm$0.11
 & 1.41$\pm$0.04
\\
& A6000 (s) $\downarrow$
 & \textit{not supported} 
 & 2.08$\pm$0.10
 & 11.49$\pm$0.20
 & 1.25$\pm$0.05
 & 1.99$\pm$0.09
 & 12.10$\pm$0.08
 & 3.09$\pm$0.02
 & 12.57$\pm$0.05
 & \textbf{0.88$\pm$0.02}
\\\bottomrule
\end{tabular}}
\vspace{-0.5em}
\captionof{table}{
    Quantitative assessment of unconditional generation. We report the mean and standard deviation for each metric. While GPNN and GPDM perform best in terms of SIFID~\cite{shaham_singan_2019}, we find that they sample near-duplicates of the input image with high probability (see~\cref{sec:supp-results}). 
    Our approach performs on par with or better than other single-image diffusion models in terms of SIFID as well as NIQE, NIMA, and MUSIQ (no-reference image quality metrics)~\cite{mittal2012making,talebi2018nima,ke2021musiq}. 
    We improve over other methods in terms of the diversity of generated images (LPIPS distance and pixel diversity~\cite{kulikov_sinddm_2022}), and we avoid the long optimization times of trained methods. 
    Increasing the number of diffusion steps can improve SIFID ($T=40$, $\eta=1$), and using approximate nearest neighbours ($k=5$) accelerates inference, with only minimal quality loss (a 0.09 increase in SIFID). 
    We time the training and inference on a 186$\times$248 image, averaged over 10 runs.
    Note that the publicly available codebase for SinGAN does not support inference on A6000 GPUs.
}
\label{tab:unconditional}
\vspace{0em}
\end{minipage}
\vspace{-1.5em}
\end{figure*}

{ 
\setlength{\textfloatsep}{4pt}
\subsection{Coarse-to-Fine Image Sampling}
\vspace{-0.25em}
\label{sec:multi-scale}

We achieve coarse-to-fine image generation by first running the single-scale image sampling procedure at the coarsest image scale. 
Then, we incorporate the coarse-scale output into the generation of an image at a finer scale, and we repeat this process until we output an image at the highest-resolution scale.
We illustrate the method in Figure~\ref{fig:method}~(bottom) and provide a pseudocode description in Algorithm~\ref{alg:coarse-to-fine}.

More specifically, we first initialize a noisy image pyramid as $\{\noisypatch_{s, \timesteps}\}_{s=0}^{\numscales}\sim\normal(\mathbf{0}, \mathbf{I})$.  
Here, $s=S$ is the coarsest scale and $s=0$ corresponds to the full-resolution image.
For $s=S$, we follow the single-scale image sampling procedure exactly to sample an output image at that scale. 
For the other scales, $s \in \{S-1, \ldots, 0\}$, we follow the single-scale sampling approach to gather and denoise patches from the noisy images $\noisypatch_{s,t}$ and reconstruct a denoised image $\denoisedpatch_{s,t}$ at each diffusion timestep; 
additionally, we guide the generation process using the output $\noisypatch_{s+1, t=0}$ from the previous (coarser) scale.
The coarse-scale guidance is incorporated by applying a high-pass filter to the denoised image $\denoisedpatch_{s, t}$ at the current scale, and adding the result to an upsampled version of $\noisypatch_{s+1, t=0}$ (see the \textsc{TwoScaleBlend} function of L9 in Algorithm~\ref{alg:coarse-to-fine}). This operation can be thought of as a two-scale version of Laplacian pyramid blending~\cite{burt1983}.
Finally, we add noise to this result to compute $\noisypatch_{s, t-1}$---the noisy image at the previous diffusion timestep for this scale.
The same denoising, blending, and noise-addition steps are repeated for each scale until the output $\noisypatch_{s=0, t=0}$ at the finest scale is produced.
Note that in this coarse-to-fine sampling procedure, we denoise the image patches using a dataset of patches $\dataset_{s}$ extracted from the input image at the same scale. 

\subsection{Acceleration Techniques}
Our closed-form denoiser can be further accelerated using several complementary techniques.
First, we leverage PyTorch's fused attention kernel (based on FlashAttention~\cite{dao2022flashattention,dao2023flashattention2}), by re-formulating our patch denoiser as scaled-dot-product attention (see \cref{sec:supp-implementation}).
Additionally, we can apply our method in a compressed latent space, similar to large diffusion models~\cite{rombach2022high}.
That is, we encode the input image into a latent using a pre-trained variational auto-encoder (VAE)~\cite{labs2025flux} and then perform denoising in the latent space.
Finally, we can approximate the summation in Equation~\ref{eq:closed-form} using $k$ approximate nearest neighbors (ANN)~\cite{indyk1998approximate} found via a clustering-based index~\cite{sivic2003video}.
We use an inverted file index with $\sqrt{N}$ clusters and probe a fixed number of them at query time, which reduces the cost from $\mathcal{O}(N^2)$ to $\mathcal{O}(N^{3/2})$.
Additional implementation details are provided in \cref{sec:supp-implementation}.

\subsection{Implementation Details}
We implement our approach in PyTorch and code is publicly available on the \href{https://haojunqiu.github.io/efficient-SID/}{project webpage}.
For $T$$=$$10$ and \texttt{float32} precision, a naive PyTorch implementation of our coarse-to-fine sampling procedure takes approximately 3 seconds for a 186$\times$248 image on an NVIDIA A6000 GPU. 
We found only a modest improvement in sampled image quality for $\timesteps$$\geq$$10$ and so we use $T$$=$$10$ for all results unless otherwise stated.
For images of $\approx$ 250$\times$250 pixel resolution, we use a patch size of 15$\times$15 pixels with $\numscales$$=$$4$ scales. 
At subsequent scales, the image resolution changes by a factor of two in each dimension, and we set the number of scales so that the patch size is about half the size of the image at the coarsest scale.
We extract patches with a stride of one pixel, and we use a Gaussian with a standard deviation of $\rho$$=$$0.2$ to weight the patches before they are reassembled into an image as $\sum_{i=1}^{\numpatches} \putpatch^{(i)}_{\rho} \denoisedpatch^{(i)}$. 
To set $\sigma(\timestep)$ and $\alpha(\timestep)$, we use the flow matching schedule~\cite{lipman2023flowmatchinggenerativemodeling}, where $\alpha(t) = 1 - t/T$ and $\sigma(t) = t/T$, and we use deterministic sampling ($\eta(t) = 0$) unless otherwise specified.
In practice, we omit the two-stage blending on the $\timestep=0$ diffusion step, which we find improves the results.
}

\begin{figure*}[ht]
    \centering
    \includegraphics[width=1.0\textwidth]{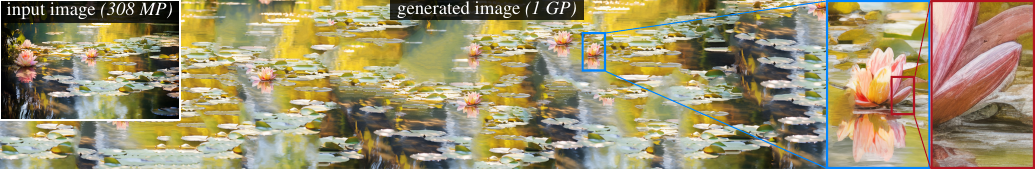}
    \caption{
        High-resolution generation. 
        The input image is 308 MP, and we generate an image of size $14336 \times 70080$ (1 GP) in only 13.9 minutes (NVIDIA RTX A6000 PRO) by incorporating the three proposed acceleration techniques (see \cref{sec:supp-high-res-implementation}).
 Specifically, we use $T = 20$ sampling steps with $\eta = 1$ and ANN with $k = 5$. \textit{Image: Duncan Rawlinson, CC BY-NC 2.0.}}
    \label{fig:high-res-gen}
    \vspace{-1.5em}
\end{figure*}

\section{Experiments}
\label{sec:experiments}

We demonstrate our approach for applications of single-image generative modeling, including unconditional generation, image retargeting, symmetrization, structural analogies~\cite{benaim2021structural}, and text-guided style transfer. 
We compare our approach to state-of-the-art single-image generative models: SinGAN~\cite{shaham_singan_2019}, SinDDM~\cite{kulikov_sinddm_2022}, SinFusion~\cite{nikankin_sinfusion_2023}, SinDiffusion~\cite{wang2022sindiffusion}, GPNN~\cite{granot_drop_2021}, and GPDM~\cite{elnekave2022generating}.
SinGAN uses a generative adversarial network, SinDDM, SinFusion, and SinDiffusion are based on diffusion models, GPNN uses patch nearest neighbors, and GPDM optimizes patch distributions via a Wasserstein distance.

\begin{table}[t]
\captionsetup{justification=raggedright,singlelinecheck=false}
\centering
\resizebox{\columnwidth}{!}{
\begin{tabular}{l|ccccccc}
\toprule
{method} 
& {256$^2$} & {512$^2$} & {1024$^2$} & {2048$^2$} & {4096$^2$} & {8192$^2$} \\
\midrule
vanilla 
& 2.27 & 42.90 & 733.75 & $>$1 hr & $>$1 hr & $>$1 hr \\
+ fused attention 
& 1.26 & 23.42 & 401.79 & $>$1 hr & $>$1 hr & $>$1 hr \\
+ latent space 
& 0.36 & 0.39 & 0.65 & 3.43 & 36.65 & 523.97 \\
+ ANN
& 0.65 & 0.96 & 1.30 & 3.85 & 15.14 & 69.39 \\

\bottomrule
\end{tabular}}
\vspace{-1em}
\caption{
Inference time versus image resolution (seconds; lower is better), measured with $T{=}10$ denoising steps on an RTX 6000 Ada (48\,GB VRAM, 12 CPUs). 
Timing covers all stages (VAE encode/decode and ANN clustering); runs are in \texttt{bfloat16} except ANN in \texttt{float32}.
\textbf{Fused attention} replaces the naive PyTorch implementation with a fused attention backend. 
\textbf{Latent space} uses the FLUX VAE~\cite{labs2025flux} with 8$\times$ spatial compression and a patch size of 7 for 16 channels.
\textbf{ANN} introduces approximate nearest neighbor search with clustering
, reducing complexity from $\mathcal{O}(N^2)$ to $\mathcal{O}(N^{3/2})$.
All methods improve efficiency.
}

\label{tab:runtime_seconds_cvpr}
\vspace{-1.5em}
\end{table}

\begin{figure}[ht]
    \centering
    \includegraphics[width=1.0\columnwidth]{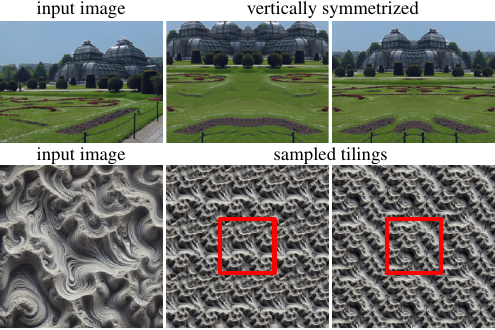}
    \vspace{-2em}
    \caption{Image symmetrization. By enforcing constraints in the diffusion process, our approach can generate new images with horizontal or vertical symmetry. It can also create images that tile together without seams (the red box indicates the sampled image, which we use to create the 3$\times$3 tiling shown).}
    \label{fig:symmetrization}
    \vspace{-1.5em}
\end{figure}

\subsection{Unconditional Single Image Generation}
\label{sec:exp-single-image-sampling}
We show examples of unconditional image sampling using our coarse-to-fine procedure in Figure~\ref{fig:unconditional}.
Qualitatively, our approach generates images with a similar appearance to the baselines, including methods that require hours of training. 
Additional qualitative results are included in~\cref{sec:supp-results}.

We provide quantitative results in Table~\ref{tab:unconditional} on the single image Fréchet Inception distance (SIFID)~\cite{shaham_singan_2019}, no-reference image quality metrics 
(NIQE, NIMA, and MUSIQ~\cite{mittal2012making,talebi2018nima,ke2021musiq}),
pixel diversity, and LPIPS diversity~\cite{kulikov_sinddm_2022}.
The latter metrics assess the diversity of generated images by measuring the average standard deviation of generated pixels across sampled images, and the average LPIPS distance between sampled image pairs.
Each metric is computed using 50 generated samples, and we report the mean and standard deviation across 15 different input images.

Our approach achieves improved SIFID compared to trained single-image diffusion models.
While GPNN and GPDM achieve the best SIFID scores, 
they often produce outputs that are nearly identical to the input image (see Figure~\ref*{fig:duplicate_comparisons} and Table~\ref*{tab:near_dups}).
We achieve comparable image quality to prior methods (NIQE, NIMA, MUSIQ), while obtaining the highest diversity among all methods (LPIPS distance, pixel diversity).
Overall, we achieve similar or better quality relative to other diffusion-based models with significantly lower computational overhead. 

\begin{figure}[ht]
    \centering
    \includegraphics[width=1.0\columnwidth]{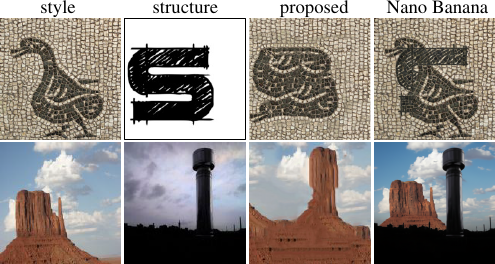}
    \vspace{-1.5em}
    \caption{Structural analogies~\cite{benaim2021structural}. Our approach combines the style of one image and the structure of another image to generate a new image that combines both properties.
        Nano Banana Pro~\cite{google2025nanobanana} preserves neither the structure nor the input patch distribution.
    }
    \label{fig:styletransfer}
    \vspace{-1.5em}
\end{figure}

In Figure~\ref{fig:high-res-gen}, we show generation of a 1 GP resolution image in 834 s by combining fused attention, latent space denoising, and approximate nearest neighbors (ANN).
Additional gigapixel examples are provided in Figure~\ref*{fig:supp-high-res}.
We find that high-resolution generation benefits from stochastic denoising ($\eta > 0$) and using ANN only at finer scales of diffusion sampling after low-spatial-frequency image components have converged (see~\cref{sec:supp-high-res-implementation}).
We assess how inference times scale with image resolution in Table~\ref{tab:runtime_seconds_cvpr}; at 16 MP resolution, the proposed acceleration techniques achieve a $>$1000$\times$ speedup relative to a naive implementation.

\subsection{Applications}
\paragraph{Image retargeting.}
We show image retargeting results in Figures~{\ref*{fig:supp-retargeting-comparison}} and ~{\ref*{fig:supp-retargeting}}.
Given a specified target resolution, our method first downsamples the input image to the resolution of the coarsest scale ($S$). 
Then, we resize the coarse-resolution image to the desired aspect ratio and use it to initialize $\hat{\mathbf{x}}_{0, S}$. 
After, we run coarse-to-fine image sampling to recover the retargeted image.
\vspace{-1em}

\paragraph{Symmetrization.}
We add constraints during the diffusion process to sample images that are vertically symmetric or that tile together without seams (see Figure~\ref{fig:symmetrization}).
We create vertical symmetry by flipping and copying one half of the denoised image to the other half of the image after every denoising timestep during the generation process.
To create tileable images, we run three separate denoising passes at every timestep. In the first pass, we directly denoise the image.
In the second and third passes, we circularly shift the image along the horizontal or vertical dimensions by half the image width or height before denoising.
After denoising, we shift the images back to their original configuration, and we blend the results of each denoising pass to ensure that the boundaries of the image are tileable.
We show results of this procedure for multiple different scenes in Figure~\ref{fig:symmetrization}; 
see~\cref{sec:supp-application,sec:supp-results} for additional details and results.
\vspace{-1em}

\paragraph{Structural analogies.}
In Figure~\ref{fig:styletransfer}, we show images created by structural analogy, which seeks to  apply the ``style'' of one image to the ``structure'' of another image.
Large diffusion models struggle with this task and fail to preserve the distribution of patches from the style image (e.g., Nano Banana Pro~\cite{google2025nanobanana}; see Sec.~\ref*{sec:supp-comparisons-to-large-diffusion} for implementation details). 

This task is accomplished by first downsampling the structure image to a coarse scale and running the single-scale DDIM inversion~\cite{dhariwal2021diffusion,mokady2023null,zhang2025geometry} to a timestep $t'{=}T/10$ or $T/2$
(we use less noise for images with high-frequency structure).
DDIM inversion converts a clean image into a corresponding noisy latent by applying 
\cref{eq:DDIM-step-1,eq:DDIM-step-2,eq:DDIM-step-3,eq:DDIM-step-4} in reverse time:
at each step, we use the predicted $\denoisedpatch_t$ and $\hat{\boldsymbol{\epsilon}}_t$ 
to deterministically produce a \emph{noisier} sample 
$\noisypatch_{t+1} \gets \alpha(t{+}1) \denoisedpatch_{t} + \sigma(t{+}1) \hat{\boldsymbol{\epsilon}}_{t}$ with $\eta = 0$.
Iterating the clean image from $t=0$ to $t=t'$ yields an inverted noisy image $\noisypatch_{t'}$ that preserves the spatial structure of the original image, which is useful for editing tasks (see \cref{sec:supp-application}).
Then, the inverted noisy image at the coarsest scale $S$ is used to initialize ${\mathbf{x}}_{t', S}$ for the coarse-to-fine image sampling procedure using the patches from the style image for denoising.
\vspace{-1em}

\paragraph{Text-guided style transfer.}
We combine our approach with pre-trained vision--language models to enable text-guided style transfer. 
We use the CLIP ViT-B/32 model~\cite{radford2021learning} and a procedure similar to that of SinDDM for this task~\cite{kulikov_sinddm_2022}.

To start, we compute a noisy version of the input image via DDIM inversion,
which we then use to initialize the reverse-diffusion process with the single-scale sampler (Algorithm~\ref{alg:single-scale-sampling}) and CLIP-guided updates. 
Specifically, at each denoising step, we first compute the denoised image $\denoisedpatch_{t}$, and the CLIP update rule is given as 
\vspace{-0.5em}
\begin{equation}
    \denoisedpatch_{t,\text{CLIP}} \leftarrow \gamma \nabla_{\denoisedpatch_{t}}\mathcal{L}_\text{CLIP} + \lambda\denoisedpatch_{t} + (1-\lambda)\denoisedpatch_{t+1,\text{CLIP}}, 
\vspace{-0.5em}
\end{equation}
where $\gamma$ is a parameter that controls the intensity of the CLIP guidance, and $\lambda$ is a momentum parameter that controls how much content to retain from the previous timestep after CLIP guidance. 
Using momentum helps to prevent image manipulations from being overridden by the denoising step~\cite{kulikov_sinddm_2022}. 
The CLIP loss $\mathcal{L}_\text{CLIP}$ is the average cosine distance between a set of augmented text and image embeddings that we compute on the input image (see~\cref{sec:supp-application}). 

We show examples of text-guided style transfer in Figures~\ref{fig:teaser} and~\ref*{fig:supp-text-style-transfer}.
The proposed approach can be adapted to generate images corresponding to various styles and artists, and we show comparisons to SinDDM in~\cref{sec:supp-results}.

\begin{figure}
\begin{minipage}{\columnwidth}
    \includegraphics{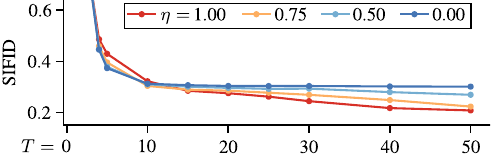}
    \vspace{-2em}
    \captionof{figure}{Plot of SIFID vs diffusion timesteps $T$ for coarse-to-fine image sampling across different $\eta$ values. The SIFID converges in roughly 10 timesteps.}
    \label{fig:steps}
    \vspace{0.5em}
\end{minipage}

\begin{minipage}{\columnwidth}
\centering
\resizebox{\columnwidth}{!}{
\begin{tabular}{cccccc}
    \toprule
    \multirow{2}{*}{\makecell{patch \\size}}& \multicolumn{5}{c}{$\rho$} \\
     & $\infty$ & 3.0 & 1.0 & 0.2 & 0.1 \\\midrule
     5 & \cellcolor[rgb]{0.598,0.698,0.778} 2.2 $\pm$ 0.9 & \cellcolor[rgb]{0.598,0.698,0.778} 2.2 $\pm$ 0.9 & \cellcolor[rgb]{0.590,0.717,0.779} 2.4 $\pm$ 1.2 & \cellcolor[rgb]{0.566,0.778,0.776} 2.7 $\pm$ 1.9 & \cellcolor[rgb]{0.562,0.789,0.774} 2.8 $\pm$ 2.0 \\
         
     7 & \cellcolor[rgb]{0.570,0.767,0.778} 2.7 $\pm$ 1.4 & \cellcolor[rgb]{0.571,0.763,0.778} 2.6 $\pm$ 1.3 & \cellcolor[rgb]{0.608,0.676,0.775} 2.1 $\pm$ 1.0 & \cellcolor[rgb]{0.629,0.626,0.762} 1.9 $\pm$ 1.0 & \cellcolor[rgb]{0.635,0.609,0.755} 1.8 $\pm$ 1.2 \\
         
     9 & \cellcolor[rgb]{0.577,0.840,0.752} 3.1 $\pm$ 1.8 & \cellcolor[rgb]{0.573,0.837,0.754} 3.1 $\pm$ 1.8 & \cellcolor[rgb]{0.596,0.702,0.778} 2.3 $\pm$ 1.3 & \cellcolor[rgb]{0.640,0.580,0.736} 1.6 $\pm$ 0.9 & \cellcolor[rgb]{0.639,0.531,0.693} 1.4 $\pm$ 0.8 \\
         
    11 & \cellcolor[rgb]{0.689,0.896,0.689} 3.5 $\pm$ 2.1 & \cellcolor[rgb]{0.676,0.892,0.696} 3.5 $\pm$ 2.1 & \cellcolor[rgb]{0.583,0.732,0.779} 2.4 $\pm$ 1.6 & \cellcolor[rgb]{0.641,0.568,0.727} 1.6 $\pm$ 0.9 & \cellcolor[rgb]{0.634,0.502,0.665} 1.3 $\pm$ 0.8 \\
         
    15 & \cellcolor[rgb]{0.886,0.939,0.566} 4.0 $\pm$ 2.9 & \cellcolor[rgb]{0.865,0.936,0.578} 3.9 $\pm$ 2.9 & \cellcolor[rgb]{0.564,0.782,0.776} 2.8 $\pm$ 2.1 & \cellcolor[rgb]{0.641,0.578,0.735} 1.6 $\pm$ 1.3 & \cellcolor[rgb]{0.635,0.507,0.671} 1.3 $\pm$ 1.1 \\
         
    19 & \cellcolor[rgb]{0.978,0.951,0.559} 4.2 $\pm$ 3.5 & \cellcolor[rgb]{0.963,0.949,0.552} 4.1 $\pm$ 3.5 & \cellcolor[rgb]{0.562,0.820,0.764} 3.0 $\pm$ 2.8 & \cellcolor[rgb]{0.633,0.614,0.757} 1.8 $\pm$ 1.8 & \cellcolor[rgb]{0.640,0.539,0.701} 1.5 $\pm$ 1.5 \\
         
    23 & \cellcolor[rgb]{0.997,0.953,0.572} 4.2 $\pm$ 4.1 & \cellcolor[rgb]{0.987,0.952,0.565} 4.2 $\pm$ 4.0 & \cellcolor[rgb]{0.583,0.845,0.748} 3.1 $\pm$ 3.2 & \cellcolor[rgb]{0.624,0.639,0.767} 1.9 $\pm$ 2.3 & \cellcolor[rgb]{0.642,0.563,0.722} 1.6 $\pm$ 1.9 \\
\end{tabular}}
\captionof{table}{Analysis of the single-scale image sampling SIFID versus different patch sizes and values of $\rho$ (used in the operator $\putpatch_\rho^{(i)}$ to assemble an image from patches). We find that smaller values of $\rho$ and patch sizes of around 11--15 pixels achieve the lowest SIFID.}
\label{tab:patch-size}
\end{minipage}
\vspace{-1.5em}
\end{figure}

\subsection{Analysis of Hyperparameter Settings}
We evaluate the sensitivity of the method to hyperparameters, including the number of diffusion timesteps, the patch size, and the value of $\rho$ (used to reassemble the image from patches $\putpatch_\rho^{(i)}$), and we report the effect on the SIFID score.
We compute metrics using the same 15 images as Table~\ref{tab:unconditional}, but for computational expediency we report the mean and standard deviation over five generated samples per image.

Figure~\ref{fig:steps} plots the SIFID vs.\ diffusion timesteps for coarse-to-fine image sampling across different $\eta$ values. 
While increasing the number of diffusion timesteps improves the SIFID scores, there are diminishing returns for $T$$>$$10$. 
Higher $\eta$ values yield larger improvements in SIFID as $T$ increases, but at the cost of reduced sample diversity (see Table~\ref{tab:unconditional}).
We also plot the impact of patch size and $\rho$ on SIFID in Table~\ref{tab:patch-size} for single-scale image sampling. 
Based on these results, we choose a patch size of 15 and $\rho=\text{0.2}$ for all our experiments.

\vspace{-0.5em}
\section{Concluding Remarks}
\label{sec:conclusion}
Recent diffusion models are trained on increasingly large datasets and have prohibitive computational costs.
Counter to this trend, we explore reducing the size of the training dataset to the bare minimum---a single image.
In this setting, closed-form denoising eliminates the need for hours of training and makes single-image generative modeling significantly more practical.
Our work opens up multiple exciting directions: we envision improving the efficiency further to enable real-time generation, developing off-the-shelf single-image priors for solving inverse problems~\cite{chung_diffusion_2023}, and introducing new diffusion priors that leverage the internal structure of multiple images simultaneously.
We discuss these extensions in more detail in Sec.~\ref*{sec:supp-discussion-and-extentions}.

\paragraph{Acknowledgments.} DBL and KNK acknowledge support of NSERC under the RGPIN program. DBL also acknowledges support from the Canada Foundation for Innovation and the Ontario Research Fund.

{
    \small
    \bibliographystyle{ieeenat_fullname}
    \bibliography{main}
}

\clearpage
\setcounter{page}{1}
\addtocontents{toc}{\protect\setcounter{tocdepth}{2}}  

\setcounter{section}{0}
\renewcommand{\thesection}{S\arabic{section}}
\renewcommand{\theHsection}{S\arabic{section}}
\setcounter{figure}{0}
\renewcommand{\thefigure}{S\arabic{figure}}
\renewcommand{\theHfigure}{S\arabic{figure}}
\setcounter{table}{0}
\renewcommand{\thetable}{S\arabic{table}}
\renewcommand{\theHtable}{S\arabic{table}}
\setcounter{equation}{0}
\renewcommand{\theequation}{S\arabic{equation}} 
\renewcommand{\theHequation}{S\arabic{equation}}
\setcounter{algorithm}{0}
\renewcommand{\thealgorithm}{S\arabic{algorithm}}
\renewcommand{\theHalgorithm}{S\arabic{algorithm}}

\onecolumnnow
\begin{center}
    \Large{\textbf{Efficient and Training-Free Single-Image Diffusion Models}\\\vspace{0.5em} Supplementary Material}
\end{center}
\tableofcontents

\section{Connections to Prior Single-Image Generative Methods}
\label{sec:supp-connections-to-other-sin-gen}

In this section, we discuss the connections and differences between our method and prior single-image generative approaches.
    Our method is a multi-scale framework that iteratively refines image patches at each scale, drawing inspiration from SinGAN~\cite{shaham_singan_2019}, GPNN~\cite{granot_drop_2021}, GPDM~\cite{elnekave2022generating}, and SinDDM~\cite{kulikov_sinddm_2022}.
As in these works, patches at coarser scales govern the global structure of the generated image while patches at finer scales control its texture.
Additionally, iterative patch refinement is performed at each scale, similar to previous work. 
However, our approach also differs from each of these in key ways, as we describe in the following sections.

\subsection{Comparison with GPNN}
Both GPNN \cite{granot_drop_2021} and our method employ coarse-to-fine sampling, 
iteratively refining patches and reassembling them into an image.
GPNN only injects noise once at the coarsest-scale initialization and then performs hard patch replacement with the (single) nearest neighbor at each iteration.
In contrast, we use diffusion as the patch-selection mechanism, which employs an optimal diffusion denoiser that operates via a weighted average of all patches, and also injects noise at every sampling step.
As the optimal denoiser approaches the final step (i.e., $\sigma(t) \to 0$), we also select or copy a single clean patch; however, all previous intermediate iterations of soft patch selection ground the probabilistic sampling for the patch prior $p_{\text{data}}$.
The one-time mild noise injection in GPNN causes samples to frequently duplicate much of the reference image---behavior that we do not observe in our approach. 
We provide an analysis of this effect in \cref{sec:supp-results}, Figure~{\ref{fig:duplicate_comparisons}}, and Table~{\ref{tab:near_dups}}.

\subsection{Comparison with GPDM}
GPDM \cite{elnekave2022generating} has a similar multi-scale formulation as GPNN, 
but it uses a very different patch generation mechanism based on an iterative sliced Wasserstein distance (SWD) optimization.
Specifically, GPDM aims to optimize the Wasserstein distance between the generated patch distribution and the reference patch distribution, where SWD provides a tractable approximation to the Wasserstein distance.
Our diffusion formulation shares the same goal, namely to sample a patch distribution that matches that of the reference.
However, diffusion naturally enables the incorporation of various guidance and sampling techniques, which may not be straightforward to implement within the SWD optimization framework.

\subsection{Comparison with SinDDM}

\begin{figure}[t]
  \centering
   \includegraphics{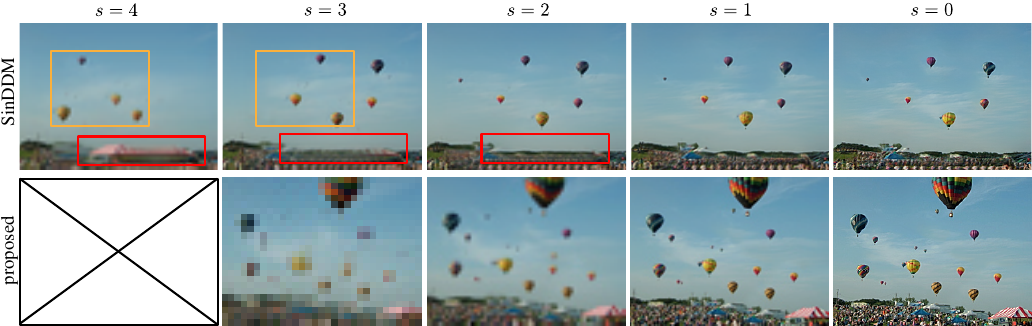}
   \caption{Multi-scale consistency comparison vs.\ SinDDM. SinDDM outputs (top) show significant inconsistency across scales, while our method (bottom) produces consistent outputs across scales.}
  \label{fig:sinddm_issue}
  \vspace{-1.25em}
\end{figure}

SinDDM \cite{kulikov_sinddm_2022} is a multi-scale trained diffusion model that performs patch-level generation by restricting the receptive field of a convolutional neural network (CNN).
The most notable difference in our method is that it does not require training, but instead uses the closed-form denoiser.
In addition, the exact multi-scale formulation also differs in the following ways. 
(1) We use a forward process with no explicit blurring (only noising), and 
(2) we use Laplacian blending across scales instead of scheduled linear interpolation to better preserve coarse-scale structure (e.g., compare Algorithm~\ref{alg:coarse-to-fine}, L9,16--17 to SinDDM Sec.~3.1, Algorithm~2, L8). 
Qualitatively, our outputs across coarse scales are more consistent than those of SinDDM (Figure~{\ref{fig:sinddm_issue}}). 
Consistency across scales is a very important property for applications such as structural analogies, where the structure image is only injected at the coarsest scale.

\newpage
\section{Closed-Form Diffusion Denoiser}
\label{sec:supp-closed-form}

\subsection{Derivation of the Closed-Form Diffusion Denoiser}
\label{sec:supp-closed-form-derivation}
It is known that the diffusion denoiser or the score estimator can be computed in closed form when having access to the empirical distribution, i.e., the real dataset \cite{karras_elucidating_2022, scarvelis_closed-form_2023}. Here, we adopt a similar derivation approach as outlined in Appendix B.3 of \cite{karras_elucidating_2022}, with adjustments tailored to our specific objective function and the formulation of diffusion models.

Let us recall that the data distribution is defined as $p_{\text{data}}(\vy) = \frac{1}{Y}\sum_{j=1}^{Y} \delta\left(\vy - \vy^{(j)}\right)$ for a dataset $\dataset = \left\{\vy^{(1)}, \dots,  \vy^{(Y)}\right\} \subseteq \R^{d}$, and the forward perturbation process of $\noisypatch_t = \alpha_t \patch + \sigma_{t} \boldsymbol{\epsilon}$, $\boldsymbol{\epsilon} \sim \mathcal{N}(\mathbf{0}, \mathbf{I})$ when given a specific clean signal $\vy$, \ie, $p(\vx_{t} \,|\, \vy) = \mathcal{N}\left(\vx_{t}; \alpha_t \vy, \sigma_t^2 \mathbf{I}\right)$. We can derive that the marginal distribution $p_{t}(\vx_{t})$ of perturbed data at some $t \in [0, T]$ is a mixture of Gaussians. 
\begin{align}
    p_t(\vx_{t}) 
    &= \int_{\R^{d}}
    p(\vx_t, \vy)
    \,\mathrm{d}\vy \\
    &= 
    \int_{\R^{d}}
    p(\vx_{t} \,|\, \vy) \,
    p_{\text{data}}(\vy) 
    \,\mathrm{d}\vy \\
    &= 
    \int_{\R^{d}}
    p(\vx_{t} \,|\, \vy) \,
    \left(\frac{1}{Y}
    \sum_{j=1}^{Y} 
    \delta(\vy - \patch^{(i)}) \right) 
    \,\mathrm{d}\vy
    \\
    &= \frac{1}{Y}
    \sum_{j=1}^{Y} 
    \int_{\R^{d}}
        p(\vx_{t} \,|\, \vy) 
        \delta(\vx - \patch^{(i)})
    \,\mathrm{d}\vy \\
    &= 
    \frac{1}{Y}
    \sum_{j=1}^{Y} 
    p(\vx_{t} \,|\, \vy = \vy^{(j)}) \\
    &= 
    \frac{1}{Y} \sum_{j=1}^{Y} 
    \mathcal{N}(\vx_{t}; \alpha_t \patch^{(i)}, \sigma_{t}^{2} \mathbf{I}),
\end{align}
Then, recall that the training objective at a specific $t$ is
\begin{align}
    \mathcal{L}(\denoiser; t) 
    &=  \mathbb{E}_{
        \vy \sim \dataset, 
        \boldsymbol{\epsilon} \sim \mathcal{N}(\mathbf{0}, \mathbf{I})}
        \left[  
            w_t
            \|
            \denoiser(
                \alpha_t \vy + \sigma_t \boldsymbol{\epsilon}, 
                t
            )
            - \vy
            \|_{2}^{2}
        \right] \\
    &= 
    \frac{1}{Y} \sum_{j=1}^{Y}
    \mathbb{E}_{
        \boldsymbol{\epsilon} \sim \mathcal{N}(\mathbf{0}, \mathbf{I})}
        \left[  
            w_t
            \|
            \denoiser(
                \alpha_t \vy^{(j)} + \sigma_t \boldsymbol{\epsilon}, 
                t
            )
            - \vy^{(j)} 
            \|_{2}^{2}
        \right] \\
    &= 
    \frac{1}{Y} \sum_{j=1}^{Y}
    \mathbb{E}_{
        \vx_{t} \sim
        \mathcal{N}(\vx_{t}; \alpha_{t} \vy^{(j)}, \sigma_{t}^{2} \mathbf{I}) }
        \left[  
            w_t
            \|
            \denoiser(
                \vx_{t}, 
                t
            )
            - \vy^{(j)}
            \|_{2}^{2}
        \right] \\
    &= \frac{1}{Y}
    \sum_{j=1}^{Y} 
    \int_{\R^{d}}
        \mathcal{N}(\vx_{t}; \alpha_t \vy^{(j)}, \sigma_{t}^2 \mathbf{I})
        w_{t} 
        \|
            \denoiser(\vx_{t}, t) - \vy^{(j)} 
        \|_{2}^{2}
    \, \mathrm{d}\vx_{t} \\
    &=
    \int_{\R^{d}}
    \frac{w_{t}}{Y}
    \sum_{j=1}^{Y}
        \mathcal{N}(\vx_{t}; \alpha_t \vy^{(j)}, \sigma_{t}^2 \mathbf{I})
            \|
                \denoiser(\vx_{t}, t) - \vy^{(j)} 
            \|_{2}^{2}
    \, \mathrm{d}\vx_{t},
\end{align}
which minimizes the loss for all possible input $\vx_{t}$ for the denoiser. When given a specific input $\vx_{t}$, the loss is 
\begin{equation}
    \mathcal{L}(\denoiser; \vx_{t}, t) 
    = \frac{w_{t}}{Y}
    \sum_{j=1}^{Y}
        \mathcal{N}(\vx_{t}; \alpha_t \vy^{(j)}, \sigma_{t}^2 \mathbf{I})
            \|
                \denoiser(\vx_{t}, t) - \vy^{(j)} 
            \|_{2}^{2}.
\end{equation}
To find the denoiser that produces the minimum mean squared error estimate, we set the gradient of the loss with respect to the denoiser to be $\mathbf{0}$:
\begin{align}
    \mathbf{0} 
    &= \nabla_{\denoiser} \mathcal{L}(\denoiser; \vx_{t}, t) \\
    &= \nabla_{\denoiser} 
    \frac{w_{t} }{Y}
    \sum_{j=1}^{Y}
        \mathcal{N}(\vx_{t}; 
        \alpha_t \vy^{(j)}, \sigma_{t}^2 \mathbf{I})
            \|
                \denoiser(\vx_{t}, t) - \vy^{(j)} 
            \|_{2}^{2} \\
    &= 
    \frac{w_{t} }{Y}
    \sum_{j=1}^{Y}
        \mathcal{N}(\vx_{t}; 
        \alpha_t \vy^{(j)}, \sigma_{t}^2 \mathbf{I})
            \nabla_{\denoiser}
            \|
                \denoiser(\vx_{t}, t) - \vy^{(j)} 
            \|_{2}^{2} \\
    &= \frac{2 w_{t}}{Y}
    \sum_{j=1}^{Y}
        \mathcal{N}(\vx_{t}; 
        \alpha_t \vy^{(j)}, \sigma_{t}^2 \mathbf{I})
            \left(\denoiser(\vx_{t}, t) - \vy^{(j)} \right) \\
    &= \frac{2 w_{t}}{Y}
    \left(
        \sum_{j=1}^{Y}
        \mathcal{N}(\vx_{t}; 
        \alpha_t \vy^{(j)}, \sigma_{t}^2 \mathbf{I})
            \;
            \denoiser(\vx_{t}, t)
        \;-\; 
        \sum_{j=1}^{Y}
            \mathcal{N}(
            \vx_{t}; 
            \alpha_t 
            \vy^{(j)}, 
            \sigma_{t}^2 \mathbf{I}) \;
            \vy^{(j)}
    \right).
\end{align}
Then, we rearrange the equation to obtain that the denoiser that minimizes the loss at the noise level $t$, given a specific noisy input $\vx_{t}$,  is
\begin{align}
    \denoiser(\vx_t, t) 
    \;=\; 
    \frac
    {
        \sum_{j=1}^{Y} \mathcal{N}
        \left(\vx_t; \alpha_t \vy^{(j)}, 
        \sigma_{t}^{2} \mathbf{I}
        \right)
        \, \vy^{(j)}
    }
    {
        \sum_{j=1}^{Y} \mathcal{N}
        \left(\vx_t; \alpha_t \vy^{(j)}, \sigma_{t}^{2} \mathbf{I}
        \right)
    }
    \;=\;
    \frac
    {
        \sum_{j=1}^{Y} \exp\left(-\frac{1}{2\sigma_t ^2} \left\| \mathbf{x}_t - \alpha_t \mathbf{y}^{(j)} \right\|_2^2\right)
        \, \vy^{(j)}
    }
    {
        \sum_{j=1}^{Y} \exp\left(-\frac{1}{2\sigma_t ^2} \left\| \mathbf{x}_t - \alpha_t \mathbf{y}^{(j)} \right\|_2^2\right)
    }.
\end{align}
Several other interpretations of this formula include kernel regression:
\begin{align}
    D(\noisypatch_t, t) =  
    \frac{
        \sum_{j=1}^{Y} k_{\sigma_t}\left(\vx_t, \alpha_t \vy^{(j)}\right) \, \vy^{(j)}
    }
    {
        \sum_{j=1}^{Y} k_{\sigma_t}\left(\vx_t, \alpha_t \vy^{(j)}\right)
    }, \label{eq:denoiser_as_kernel}
\end{align}
where $k_{\sigma}(\vx, \vy) = \exp\left( -\frac{\|\vx - \vy \|_{2}^{2} }{2 \sigma^{2}} \right)$ denotes the Gaussian kernel with bandwidth $2\sigma^2$.
Another equivalent interpretation is as the empirical posterior mean based on the finite data samples:
\begin{align}
    \denoiser(\vx_{t}, t) 
    &= 
    \frac
    {
        \sum_{j=1}^{Y} 
        p(\vx_{t} \,|\, \vy = \vy^{(j)})\, \vy^{(j)}
    }
    {
        \sum_{j=1}^{Y} 
        p(\vx_{t} \,|\, \vy = \vy^{(j)})
    } \\
    &= 
    \sum_{j=1}^{Y} 
    \frac
    {
        p(\vx_{t} \,|\, \vy = \vy^{(j)})
    }
    {
        \sum_{j'=1}^{Y} 
        p(\vx_{t} \,|\, \vy = \vy^{(j')})
    }
    \vy^{(j)} \\
    &= 
    \sum_{j=1}^{Y} 
    \frac
    {
        p(\vx_{t} \,|\, \vy = \vy^{(j)}) \,
        p(\vy = \vy^{(j)})
    }
    {
        \sum_{j'=1}^{Y} 
        p(\vx_{t} \,|\, \vy = \vy^{(j')}) \, p(\vy = \vy^{(j')})
    }
    \vy^{(j)} \\
    &= 
    \sum_{j=1}^{Y} 
    \frac
    {
        p(\vx_{t}, \vy = \vy^{(j)})
    }
    {
        p(\vx_{t})
    }
    \vy^{(j)} \\
    &= \mathbb{E}_{\mathbf{\vy} \sim p(\vy \,|\, \vx_t)} [\vy \,|\, \vx_t]
    \, \label{eq:emprical_posterior_mean}.
\end{align}
Finally, we can introduce a softmax notation:
\begin{align}
    \denoiser(\noisypatch_t, t) 
    = \sum_{j=1}^{Y} 
    \frac
    {\exp \left(-\frac{1}{2\sigma_t^2} \left\| \noisypatch_t - \alpha_t \patch^{(j)} \right\|_2^2 \right)}
    {\sum_{j'=1}^{Y} 
    \exp \left(-\frac{1}{2\sigma_t^2} \left\| \noisypatch_t - \alpha_t \patch^{(j')} \right\|_2^2 \right)
    } 
    \patch^{(j)} 
    &= \sum_{j=1}^{Y}
    \operatorname{softmax}
    \left( \left[ -\frac{1}{2\sigma_t^2} \left\| \noisypatch_t - \alpha_t \patch^{(j')} \right\|_2^2 \right]_{j'} \right)_j \patch^{(j)} 
\end{align}
where $\left[ -\frac{1}{2\sigma_t^2} \left\| \noisypatch_t - \alpha_t \patch^{(j')} \right\|_2^2 \right]_{j'}$ 
denotes a $Y$-dimensional vector. We will see in \cref{subsec:supp-implementation-flashattn} that this helps connect the closed-form denoiser to the attention operation.

\subsection{Computability and Generalization of Closed-Form Denoisers}
Closed-form denoisers
\cite{karras_elucidating_2022,scarvelis_closed-form_2023,niedoba2025mechanisticexplanationdiffusionmodel,kamb2025an,bertrand2025on,buchanan2025edgememorizationdiffusionmodels,lukoianov2025localityimagediffusionmodels,song2025selectiveunderfittingdiffusionmodels}
have been extensively analyzed, but so far have seen limited practical use.  
Two challenges are commonly raised: \emph{computability} and generalization/memorization.
Our setting provides a sweet spot where both concerns are alleviated.

First, the full closed-form denoiser is computationally infeasible when applied
to an entire internet-level image dataset, where each data point is itself a very high-dimensional
vector and the summation involves millions of terms.  
In contrast, our method operates at the \emph{patch level} of a \emph{single}
image, where each patch is small and the dataset consists of
a finite number of patches.  With the acceleration techniques introduced in this paper,
closed-form denoising becomes practical and efficient.

Second, closed-form denoisers encode the prior as a sum of Dirac delta
densities, so when $\sigma_t \to 0$, the distribution collapses onto a single
term, which can raise concerns about memorization or overfitting when applied
at the full-image level.
In our case, however, the denoiser operates on \emph{patches} rather than
entire images, and our image sampler stitches these patchwise predictions into
a rich and diverse distribution over global images.  
This is supported by both our diversity metrics (Table~\ref{tab:unconditional})
and the unconditional generations shown in Figure~\ref{fig:supp-unconditional}.

Our observations connect naturally to the findings of Niedoba et al.~\cite{niedoba2025mechanisticexplanationdiffusionmodel} 
and Kamb et al.
\cite{kamb2025an}, which argue that
neural network-based diffusion models generalize by performing locally optimal denoising
operations across the training distribution.  
In contrast to their dataset-level analysis, we study this behavior in the
\emph{single-image} regime, in the spirit of the SinGAN and SinDDM literature,
and show that patch-level closed-form denoising can be used to support a wide
range of practical single-image applications, including unconditional generation, 
text-guided stylization, structural analogy, symmetrization and retargeting.

\newpage
\section{Supplementary Implementation Details}
\label{sec:supp-implementation}
In this section, we provide additional implementation details and describe several complementary techniques used to accelerate the patch-based closed-form denoiser. Specifically,
\cref{subsec:supp-implementation-naive} outlines how we batch-denoise patches using straightforward PyTorch operations;
\cref{subsec:supp-implementation-flashattn} shows how closed-form denoising maps onto the attention operation, enabling the use of fused kernels such as FlashAttention~\cite{dao2022flashattention,dao2023flashattention2,shah2024flashattention} for substantial speedups;
\cref{subsec:supp-implementation-vae} explains how FLUX’s variational autoencoder (VAE)~\cite{labs2025flux} spatially compresses RGB images into latent representations, providing a large constant-factor acceleration;
\cref{subsec:supp-implementation-ann} describes how approximate nearest-neighbour search further improves asymptotic runtime; and
\cref{sec:supp-high-res-implementation} details how all of these techniques combine to enable efficient generation of gigapixel-resolution images.

\begin{figure*}[h]
    \centering
    \includegraphics[width=\textwidth]{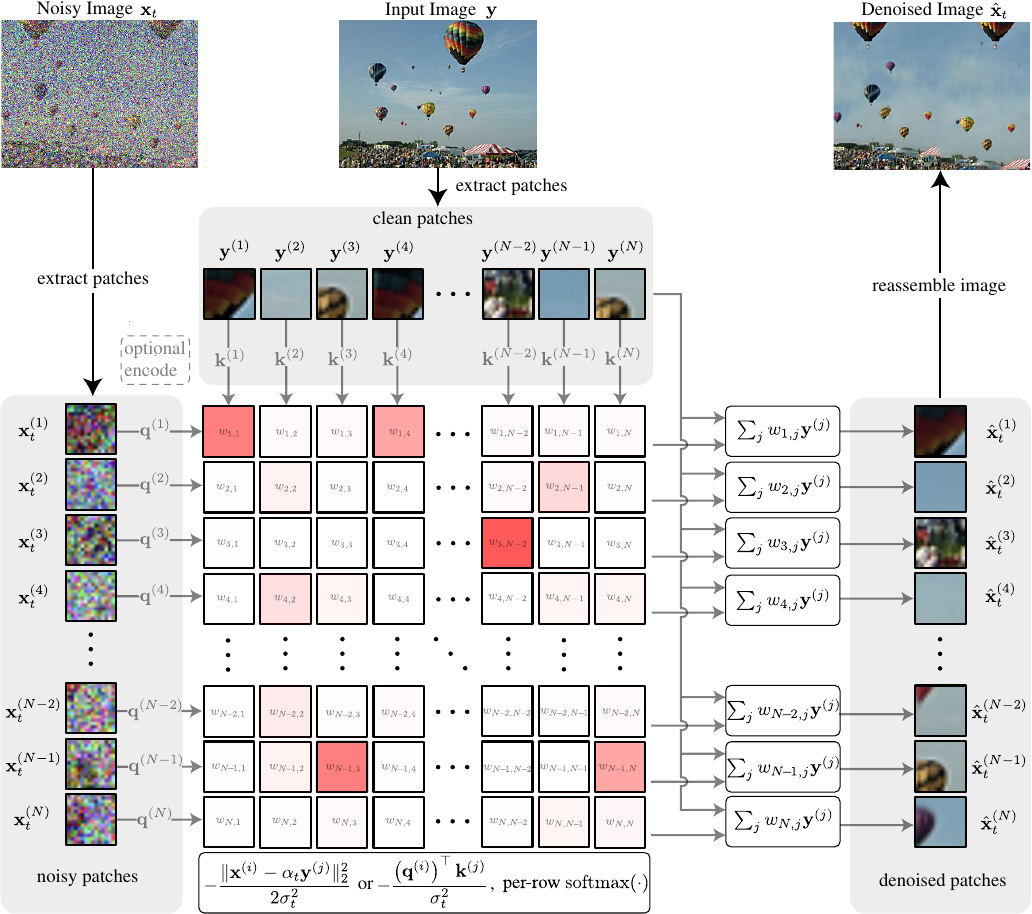}
    \caption{
        Overview of the batched closed-form denoising implementation.
        This figure illustrates both the naive vectorized PyTorch implementation 
        in \cref{subsec:supp-implementation-naive} and the version in 
        \cref{subsec:supp-implementation-flashattn} that implements this operation 
        using attention. 
        We assume the number of clean and noisy patches is the same ($Y = N$).
        To denoise a noisy image from $\denoisedpatch_t$ to $\denoisedpatch$, 
        we extract both the noisy patches $\mathbf{x}_t^{(i)}$ and clean patches $\mathbf{y}^{(j)}$ 
        and compute a weight matrix whose $(i,j)$-th entry $w_{i,j}$ predicts 
        each denoised patch as a weighted sum of the clean patches.
        Finally, we assemble the denoised patches back to their original positions 
        to obtain the denoised image.
    }
    \label{fig:supp-patch-denoise-attention}
\end{figure*}

\subsection{Naive Closed Form Denoiser Implementation}
\label{subsec:supp-implementation-naive}

\paragraph{Vectorization.} \cref{sec:supp-closed-form} gives an overview of closed-form denoising of a single data sample. 
In a single diffusion sampling step, given a noisy image $\vx_t$, we extract a batch of noisy patches ${\noisypatch_t^{(i)}}$ to be denoised.
Because these patches are relatively low dimensional, we can denoise them in parallel and fully leverage GPU acceleration.
An illustration of the batched computation appears in Figure~\ref{fig:supp-patch-denoise-attention}.
Specifically, we stack the denoised patches $\denoisedpatch_{t}^{(i)} = \denoiser(\noisypatch^{(i)}, t)$, the noisy patches $\noisypatch_t^{(i)}$, and the clean reference patches $\patch^{(i)}$ to form the rows of the following matrices:
\begin{align}
    \hat{\mathbf{X}}_{t}
    =
    \underbrace{
        \begin{bmatrix}
        \text{---}\; (\hat{\mathbf{x}}_{t}^{(1)})^{\top} \text{---}\\
        \text{---}\; (\hat{\mathbf{x}}_{t}^{(2)})^{\top} \text{---}\\
        \vdots \\
        \text{---}\; (\hat{\mathbf{x}}_{t}^{(N)})^{\top} \text{---}\;
        \end{bmatrix}
    }_{\in \mathbb{R}^{N \times d}},
    \quad
    \mathbf{X}_t
    =
    \underbrace{
        \begin{bmatrix}
        \text{---}\; ({\mathbf{x}}_{t}^{(1)})^{\top} \text{---}\\
        \text{---}\; ({\mathbf{x}}_{t}^{(2)})^{\top} \text{---}\\
        \vdots \\
        \text{---}\; ({\mathbf{x}}_{t}^{(N)})^{\top} \text{---}\;
        \end{bmatrix},
    }_{\in \mathbb{R}^{N \times d}}
    \quad
    \mathbf{Y}
    =
    \underbrace{
        \begin{bmatrix}
        \text{---}\; ({\mathbf{y}}^{(1)})^{\top} \text{---}\\
        \text{---}\; ({\mathbf{y}}^{(2)})^{\top} \text{---}\\
        \vdots \\
        \text{---}\; ({\mathbf{y}}^{(Y)})^{\top} \text{---}\;
        \end{bmatrix} 
    }_{\in \mathbb{R}^{Y \times d}}.
\end{align}
Then, we can write batched denoising in vectorized form as
\begin{align}
    \hat{\mathbf{X}}_t = 
    \underbrace{
        \operatorname{softmax}
        \left( \mathbf{L}(\mathbf{X}_t, \mathbf{Y}) \right) 
    }_{\mathbf{W} \in \mathbb{R}^{N \times Y}}
    \mathbf{Y},
\end{align}
where $\mathbf{L}$ is a matrix in $\mathbb{R}^{N \times Y}$ whose $(i,j)$-th entry 
is $-\frac{1}{2\sigma_t^2}\left\| \mathbf{x}_t^{(i)} - \alpha_t\mathbf{y}^{(j)} \right\|_2^2$
and matrix $\mathbf{W} = \operatorname{softmax}(\mathbf{L})$ represents row-wise normalization of $\mathbf{L}$
so that each row sums to $1$.
In PyTorch, this can be implemented using \texttt{torch.cdist} or by forming $\mathbf{L}$ via matrix multiplications, applying a row-wise softmax, and finally computing the weighted sum of clean patches through the matrix product $\mathbf{W}\mathbf{Y}$.

\paragraph{Runtime.} The asymptotic complexity of each denoising step is $\mathcal{O}(N \times Y \times d)$,
which becomes $\mathcal{O}(N^2\times d)$ when the number of clean patches and noisy patches is the same.
\subsection{Acceleration by a Fused FlashAttention-like Kernel}
\label{subsec:supp-implementation-flashattn}
\paragraph{Connections of the closed-form denoiser to attention~\cite{vaswani2017attention}.} 
An important insight to our closed-form denoiser is that it can be treated as scaled dot-product attention. 
Thus, we can benefit from using FlashAttention and potentially any other hardware-accelerated algorithm.
Observe that the Gaussian kernel in general can be decomposed as
\begin{align}
    k_{\sigma}(\vx, \vy) 
    = \exp{\left(-\frac{\| \vx - \vy \|_2^2}{2\sigma^2}\right)} 
    = \exp{\left(-\frac{\| \vx \|_2^2 -2 \vx^{\top} \vy + \|\vy\|_2^2}
        {2\sigma^2}\right)
    }
    = \exp{\left(-\frac{\| \vx \|_2^2 / 2}
        {\sigma^2} \right)} 
      \exp{\left(\frac{ \vx^{\top} \vy - \|\vy\|_2^2 / 2}
        {\sigma^2} \right)}.
\end{align}
In particular, the $(i,j)$-th entry of $\mathbf{W}$ as softmax weight 
can be simplified as follows due to the shift-invariance of softmax:
\begin{align}
    w_{ij}
    &= \frac{
        k_{\sigma_t} \left( \vx_t^{(i)}, \alpha_t \vy^{(j)} \right) 
    } 
    {
        \sum_{j'} k_{\sigma_t} \left( \vx_t^{(i)}, \alpha_t \vy^{(j')} \right)
    } \\
    &= 
    \frac{
        \exp{\left(-\frac{\left\| \vx_t^{(i)} \right\|_2^2 / 2}
        {\sigma_t^2} \right)} 
        \exp{\left(\frac{\left(\vx_t^{(i)}\right)^{\top} \left(\alpha_t \vy^{(j)}\right) - \left\|\alpha_t \vy^{(j)} \right\|_2^2 / 2 }
        {\sigma_t^2} \right)} 
    } 
    {   
        \sum_{j'}
        \exp{\left(-\frac{\left\| \vx_t^{(i)} \right\|_2^2 / 2}
        {\sigma_t^2} \right)} 
        \exp{\left(\frac{ \left(\vx_t^{(i)} \right)^{\top} \left(\alpha_t \vy^{(j')}\right) - \left\|\alpha_t \vy^{(j')} \right\|_2^2 / 2}
        {\sigma_t^2} \right)} 
    }
    \\
    &= 
    \frac{
        \exp{\left(\frac{ \left(\vx_t^{(i)}\right)^{\top} \left(\alpha_t \vy^{(j)}\right) - \left\|\alpha_t \vy^{(j)} \right\|_2^2 / 2 }
        {\sigma_t^2} \right)} 
    } 
    {   
        \sum_{j'}
        \exp{\left(\frac{ \left( \vx_t^{(i)}\right)^{\top}\left(\alpha_t \vy^{(j')}\right) - \left\|\alpha_t \vy^{(j')} \right\|_2^2 / 2}
        {\sigma_t^2} \right)} 
    }.
\end{align}
Now, using a homogeneous coordinate trick, we can express the noisy patch $\vx_{t}$ and scaled clean patch $\alpha_t \vy^{(j)}$ 
as a query and a key respectively so that it becomes a dot product inside the exponentials. Specifically,
we denote 
\begin{align}
\mathbf{q}^{(i)}
=
\begin{bmatrix}
\vx_t^{(i)} 
\vspace{0.5em}
\\
1
\end{bmatrix}, \quad 
\mathbf{k}^{(j)} =
\begin{bmatrix}
\alpha_t \vy^{(j)} \\
 -\left\| \alpha_t \vy^{(j)}\right\|_2^2 / 2 
\end{bmatrix}
\end{align}
both as $d+1$ dimensional vectors, and we have
\begin{align}
    \exp{\left(\frac{ \left(\vx_t^{(i)}\right)^{\top} \left(\alpha_t \vy^{(j)}\right) - \left\|\alpha_t \vy^{(j)}\right\|_2^2 / 2}
        {\sigma_t^2} \right)}
    = 
    \exp{\left(
        \frac{\left(\mathbf{q}^{(i)}\right)^{\top} \mathbf{k}^{(j)}}
        {\sigma_t^2}
    \right)}.
\end{align}
This suggests that once we define two more stacked matrices,
\begin{align}
    {\mathbf{Q}}
    =
    \underbrace{
        \begin{bmatrix}
        \text{---}\; ({\mathbf{q}}^{(1)})^{\top} \text{---}\\
        \text{---}\; ({\mathbf{q}}^{(2)})^{\top} \text{---}\\
        \vdots \\
        \text{---}\; ({\mathbf{q}}^{(N)})^{\top} \text{---}\;
        \end{bmatrix}
    }_{\in \mathbb{R}^{N \times (d+1)}}
    \quad \text{and} \quad
    \mathbf{K}
    =
    \underbrace{
        \begin{bmatrix}
        \text{---}\; ({\mathbf{k}}^{(1)})^{\top} \text{---}\\
        \text{---}\; ({\mathbf{k}}^{(2)})^{\top} \text{---}\\
        \vdots \\
        \text{---}\; ({\mathbf{k}}^{(Y)})^{\top} \text{---}\;
        \end{bmatrix} 
    }_{\in \mathbb{R}^{Y \times (d+1)}},
\end{align}
we can write the weight matrix as
\begin{align}
    \mathbf{W} = \operatorname{softmax}(\mathbf{L}(\mathbf{X}_t, \mathbf{Y})) = \operatorname{softmax}(\mathbf{Q}\mathbf{K}^{\top} / \sigma_t^2).
\end{align}
Then, with a dummy ``value'' $\mathbf{V} := \mathbf{Y}$, we obtain
\begin{align}
    \hat{\mathbf{X}}_{t} 
    = \operatorname{softmax}(\mathbf{L}(\mathbf{X}_t, \mathbf{Y})) \mathbf{Y}
    = \operatorname{softmax}{
        \left(\
        \frac{\mathbf{Q}\mathbf{K}^{\top}}{\sigma_t^2}
        \right)
    } \mathbf{V}
\end{align}
which is exactly the scaled dot product used in attention, with scale $1 / \sigma_t^2$ (instead of the originally used $1 / \sqrt{d}$). We can therefore utilize acceleration techniques like FlashAttention for our closed-form denoiser.

\paragraph{FlashAttention~\cite{dao2022flashattention,dao2023flashattention2,shah2024flashattention}.}
We employ the FlashAttention family of kernels~\cite{dao2022flashattention,dao2023flashattention2,shah2024flashattention} to accelerate our patch denoising operations.
FlashAttention restructures the standard scaled dot-product computation into SRAM-resident tiles and applies an online softmax update, eliminating the need to materialize the full attention matrix and thereby reducing High Bandwidth Memory (HBM) traffic and peak memory while maintaining exact outputs.

We use PyTorch's fused \texttt{scaled\_dot\_product\_attention} operator with the \emph{memory-efficient} backend, which dispatches to CUTLASS-based FlashAttention kernels.
As these kernels only support mask-like bias formats, the formulation introduced in the previous section---where the per-vector offset is absorbed via a \emph{homogeneous coordinate} trick---allows us to omit the bias tensor entirely and remain on the FlashAttention execution path.
Passing a general dense bias would otherwise route the operator to a slower fallback implementation.

To satisfy kernel alignment constraints, we pad the patch-vector dimension to architecture-friendly multiples (e.g., 16/32/64).
Padding with zeros leaves the computation unchanged while enabling more efficient vectorized memory access.
We also observe that FlashAttention loses efficiency when the patch-vector dimension becomes large (typically $d \gtrsim 1000$), so we keep all patch-vector dimensions below this range to consistently benefit from the fused kernels.

\paragraph{Runtime with FlashAttention.} The asymptotic bound is still $\mathcal{O}(N \times Y \times d)$, and $\mathcal{O}(N^2 \times d)$ when $N = Y$. 
Practically, however, the fused attention kernel provides a consistent 1.8$\times$ speedup compared to the vanilla implementation across all image resolutions, see Table~\ref{tab:runtime_seconds_cvpr}.

\subsection{Acceleration with an Image Auto-Encoder (VAE)}
\label{subsec:supp-implementation-vae}

\paragraph{Motivation.}
As in large diffusion models, attention-like computations become prohibitively expensive for high-resolution images because the cost scales quadratically with sequence length. 
If we extract one patch per pixel (stride = 1), then the number of noisy queries $N$ and clean keys $Y$ are both proportional to the number of pixels, resulting in an $\mathcal{O}(N Y)$ interaction that quickly dominates runtime.  
The latent diffusion literature alleviates this by spatially compressing the image using a VAE, reducing the sequence length while preserving perceptual quality with only mild compression loss.

\paragraph{Our VAE setup.}
We use the VAE from the FLUX model \cite{labs2025flux}, which downsamples each spatial dimension by a factor of $8$ and encodes images into a $16$-channel latent image.
This corresponds to a $1/64$ reduction in pixel count, and—when patches are extracted with stride $1$—a $1/64$ reduction in the total number of patches for both noisy and clean images.
For the same patch size, the number of channels—and hence the patch vector dimensionality—increases by a factor of $16 / 3 \approx 5.33$.
However, to keep comparisons fair, we use a latent patch size of $7$, giving a patch vector dimension
\[
d = 7 \times 7 \times 16 = 784,
\]
which is comparable to our canonical RGB patch dimension of $15 \times 15 \times 3 = 675$.
Empirically, our single-scale and multi-scale image denoisers operate just as well on latent tensors of shape $(H/8) \times (W/8) \times 16$ as on RGB images of shape $H \times W \times 3$.

\paragraph{Runtime.}
The theoretical complexity of the denoising kernel remains $\mathcal{O}(N \times Y \times d)$, or $\mathcal{O}(N^2 d)$ in the common setting $Y = N$.
However, the VAE provides a large constant-factor speedup.  
(1) The VAE encoder/decoder is convolutional and thus scales linearly with the number of patches, adding minimal overhead.  
(2) The denoising step now operates on only $(1/64)N$ noisy patches and $(1/64)Y$ clean patches, reducing the cost of the core batched patch denoising step by a factor of
\[
\left(\frac{1}{64}\right)^2 = \frac{1}{4096}
\]
relative to the RGB-space computation.  
As shown in Table~\ref{tab:runtime_seconds_cvpr}, the latent-space version yields substantial acceleration, especially at high resolutions.

\subsection{Acceleration with Approximate Nearest Neighbours}
\label{subsec:supp-implementation-ann}

The closed-form denoiser in Equation~\ref{eq:denoiser_as_kernel} can be further accelerated by restricting the kernel summation to the top-$k$ approximate nearest neighbour (ANN) patches of each query. 
We adopt the Faiss~\cite{douze2024faiss,johnson2019billion} implementation of the inverted file index~\cite{sivic2003video} for ANN search.

\paragraph{Algorithm.}
The dataset of clean patches $\mathcal{Y}$ is first partitioned into $n_{\text{list}}$ clusters using the k-means algorithm, and the resulting centroids define an inverted index. 
This index is built once and reused across all denoising steps. 
At inference time, for each noisy query patch $\vx_t$ and each diffusion timestep, we retrieve its approximate $k$ nearest neighbours by:
\begin{enumerate}
    \item finding the closest $n_{\text{probe}}$ centroids and their associated clusters;
    \item performing an exact search for the $k$ nearest neighbours within the union of those $n_{\text{probe}}$ clusters.
\end{enumerate}
The approximation comes from step~(1): if we set $n_{\text{probe}} = n_{\text{list}}$, the method reduces to exact nearest neighbour search over all patches.
In our experiments, we follow a standard choice and set $n_{\text{list}} = \sqrt{Y}$, where $Y$ is the total number of clean patches.

We optionally apply product quantization (PQ) on top of the inverted file index. 
PQ compresses each $d$-dimensional patch vector into a short code (a few bytes), which reduces memory consumption and replaces expensive distance computations with table lookups, yielding an additional constant-factor speedup in both memory and runtime.

\paragraph{Runtime.}
For simplicity, we assume $Y = N$, i.e., the number of clean patches matches the number of noisy patches. 
With $n_{\text{list}} = \sqrt{Y} = \sqrt{N}$, each cluster contains approximately $N / n_{\text{list}} = \sqrt{N}$ patches on average.
For a single query, the cost of:
(1) scanning all $n_{\text{list}} = \sqrt{N}$ centroids to find the closest $n_{\text{probe}}$ is $\mathcal{O}(\sqrt{N})$, and  
(2) searching within the $n_{\text{probe}}$ selected clusters is $\mathcal{O}(n_{\text{probe}} \sqrt{N})$.  
Thus, each query costs $\mathcal{O}((n_{\text{probe}} + 1)\sqrt{N})$, and performing this search for all $N$ noisy patches yields a total complexity of
\[
\mathcal{O}\bigl(N (n_{\text{probe}} + 1) \sqrt{N} \bigr)
= \mathcal{O}(N^{3/2})
\]
(up to constant factors and the patch dimension $d$).
This is an improvement over the $\mathcal{O}(N^2)$ cost of exact closed-form denoising, and in practice—together with PQ and GPU acceleration—provides substantial speedups while maintaining high-quality denoising.

\subsection{Implementation of High-Resolution Generation}
\label{sec:supp-high-res-implementation}

We combine all acceleration techniques described above to enable gigapixel-scale
generation with our method. 
We first encode each input image with the FLUX VAE and initialize the denoising
process in the latent space. 
All denoising operations are performed on latent tensors, and the final
high-resolution result is obtained by decoding the generated latent back to RGB
space.

For the weighted kernel summation, we use a FlashAttention-style implementation
instead of naive PyTorch operations. 
Across image scales, we employ different denoiser types for a good trade-off of efficiency and quality:
we use the exact closed-form denoiser on coarse scales whose spatial resolution
is below $500\times 500$ pixels, where exact computation is affordable; for all finer
scales, we use approximate nearest neighbours (ANN) with product quantization
(PQ).  
Our ANN configuration uses $n_{\text{probe}} = 40$, $k = 5$, and PQ codes of
28~bytes per patch vector, which significantly reduces memory footprint while
maintaining fidelity.  
We run the sampler for $T = 20$ denoising steps with full stochasticity $\eta = 1.0$.

As shown in \cref{fig:high-res-gen,fig:supp-high-res}, our implementation
produces high-quality gigapixel images on an NVIDIA RTX 6000 PRO GPU, with
end-to-end runtimes well under an hour (the three examples shown take
approximately 13, 33, and 39 minutes, respectively).

\newpage
\section{Application Details}
\label{sec:supp-application}

\subsection{Tileable Image Generation}
\begin{figure*}[t]
    \centering
    \includegraphics[width=\textwidth]{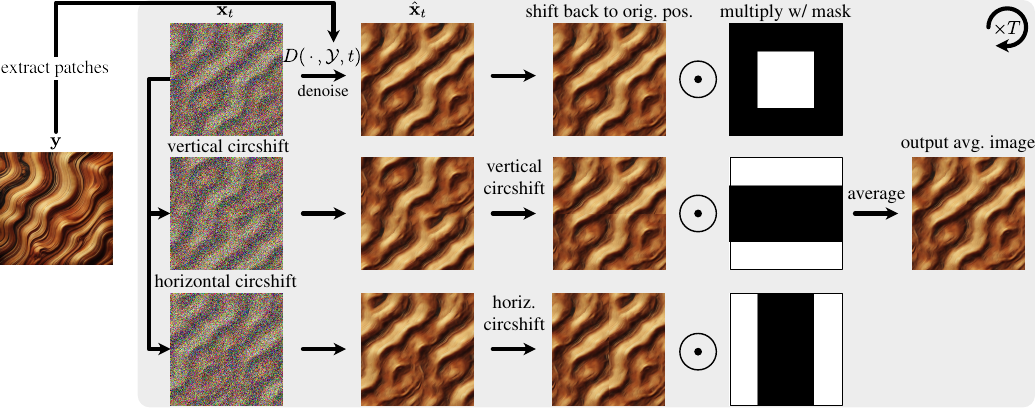}
    \vspace{-1.9em}
    \caption{Overview of tileable image generation. Starting with a noisy image $\noisypatch_t$, we circularly shift (``circshift'') this image vertically and horizontally, denoise the resulting images, and shift them back. Then, we multiply each image with a mask and average the resulting images to compute an output image. This procedure is repeated iteratively through the reverse diffusion process with the coarse-to-fine image sampling to generate tileable images.}
    \label{fig:supp-tileable-gen}
\end{figure*}

We illustrate the procedure for tileable image generation in Figure~\ref{fig:supp-tileable-gen}. 

\subsection{Single-Scale Inversion Algorithm}
\begin{algorithm}[t]\small
\caption{Single-Scale Inversion Sampling}
\label{alg:supp-inversion}
\begin{algorithmic}[1]
\Procedure{SampleImageInversion}{$\patch, t'$}
    \Comment{$t'$ is the target inversion timestep}
    \State $\dataset = \{ \getpatch^{(1)}\patch,\ldots, \getpatch^{(\numpatches)}\patch \}$ 
    \Comment{extract clean reference patches}
    \State $\noisypatch_1 \leftarrow \alpha(1) \patch + \sigma(1) \boldsymbol{\epsilon}$, \quad $\boldsymbol{\epsilon} \sim \mathcal{N}(\mathbf{0}, \mathbf{I})$  
    \Comment{first step warmup}
    \For{$\timestep = 1, \ldots, t' - 1$} 
        \Comment{clean $\rightarrow$ noisy diffusion steps} 
        \State $\denoisedpatch_{\timestep} 
            \leftarrow  \Call{ImgDenoise}{\noisypatch_\timestep, \dataset, \timestep}$
        \State $\hat{\boldsymbol\epsilon}_\timestep 
            \leftarrow \bigl(\noisypatch_\timestep - \alpha(\timestep)\,\denoisedpatch_\timestep \bigr) \big/ \sigma(\timestep)$
        \State $\noisypatch_{\timestep+1} 
            \leftarrow 
            \alpha(\timestep+1)\,\denoisedpatch_{\timestep} 
            + \sigma(\timestep+1)\,\hat{\boldsymbol \epsilon}_{\timestep}$
            \Comment{DDIM inversion step with $\eta(\timestep)=0$}
    \EndFor
    \State \Return $\noisypatch_{t'}$
\EndProcedure
\end{algorithmic}
\end{algorithm}

\begin{figure*}[ht]
    \centering
    \includegraphics[width=\textwidth]{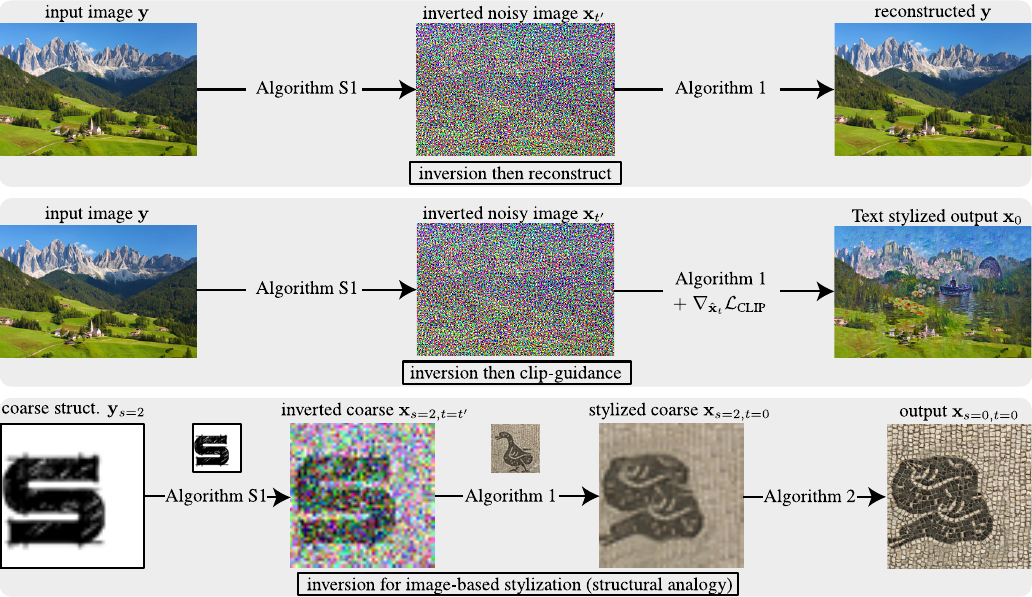}
    \caption{
    \textbf{Overview of inversion algorithms across different use cases.} %
    \textbf{Top row:} Using our inversion method (\cref{alg:supp-inversion}), an input image can be inverted to a noise level $t'$ and then reconstructed using the same patch dataset via our single-scale denoiser (\cref{alg:single-scale-sampling}). %
    \textbf{Middle row:} For text stylization, we apply the same inversion procedure to obtain the noisy image, but during denoising we additionally incorporate CLIP-guided gradients to produce a ``Monet''-style output. %
    \textbf{Bottom row:} Structural analogy. We first invert the coarse scale $s = 2$ of the structure image using its own patch dataset to obtain a coarse noisy image at $t'$. We then \emph{denoise this inverted coarse image using the style image's patch dataset} (note the patch dataset change at this stage) and finally apply coarse-to-fine sampling to produce the full-resolution output (\cref{alg:coarse-to-fine}).%
    }
    \label{fig:supp-inversion-illustration}
\end{figure*}

\paragraph{DDIM inversion with warm-start.}
To obtain a noisy version of an input image $\patch$ at a target diffusion
timestep $t'$, we use a deterministic DDIM-style inversion.  The procedure,
shown in Algorithm~\ref{alg:supp-inversion}, follows the standard DDIM update
in the forward direction (i.e., the ODE view with $\eta(t)=0$), but begins with
a small warm-start around the clean image.  Specifically, we add a small noise at the
first timestep using $(\alpha(1),\sigma(1))$ to obtain $\mathbf{x}_1$, and then apply
the usual DDIM update from $t=1$ up to $t= t'$. Here, we can invert the clean image all
the way to $t'=T$.

This warm-start is a lightweight heuristic that avoids the trivial fixed point
obtained when our closed-form denoiser is applied directly to $\mathbf{y}$.  Apart from
this first step, the trajectory is identical to standard DDIM
inversion.  Throughout the paper, we refer to this procedure simply as ``DDIM
inversion'' for brevity.

\paragraph{Optional stopping at intermediate noise levels.}
While conventional DDIM inversion typically runs to the maximal noise level
$\sigma(T)\approx 1$, our method allows stopping at any intermediate timestep
$t'$, yielding $\noisypatch_{t'}$ with noise level $\sigma(t')$.  
This provides a convenient knob that controls how much structure is preserved
before the (guided) reverse process is applied.
Lower $t'$ retains more image detail, while higher $t'$ yields stronger edits,
making this flexibility useful for style transfer and structural-analogy tasks.

\paragraph{Properties and applications.}
For any inversion depth $t'$, running our sampler
(Algorithm~\ref{alg:single-scale-sampling}) from $t'$ back to $t=0$ with the
same patch dataset reconstructs the original image up to small numerical
differences.  
The usefulness of inversion emerges when the reverse denoising is \emph{not}
a simple reconstruction but a guided or cross-image process.  
Because the inversion step preserves the coarse structure of the input image in $\mathbf{x}_{t'}$,
the subsequent guided reverse pass can introduce controlled edits while still
respecting the original layout.  
For example, replacing the reverse denoiser with a CLIP-guided version enables
text-driven stylization, where global structure is preserved and local texture
is modified.  
Similarly, in structural-analogy experiments, we invert an image using patches
from a source and denoise using patches from a target, retaining the source
layout while transferring target appearance. 
See Figure~\ref{fig:supp-inversion-illustration} for illustration of these three use cases.

\subsection{Structural Analogy}
We illustrate our structural-analogy procedure in the bottom row of Figure~\ref{fig:supp-inversion-illustration}. 
In this setting, inversion (Algorithm~\ref{alg:supp-inversion}) is performed using the patch dataset extracted from the \emph{structure} image, producing a coarse noisy image at noise level $t'$. 
During denoising (Algorithm~\ref{alg:single-scale-sampling}), we then \emph{switch to the patch dataset of the style image} to transfer its local appearance. 
Finally, we apply coarse-to-fine sampling to obtain the full-resolution output (Algorithm~\ref{alg:coarse-to-fine}).

\subsection{Text-Based Style Transfer}
The CLIP guidance used in text-based style transfer incorporates the following image and text augmentation used in Text2LIVE~\cite{bar2022text2live} and SinDDM~\cite{kulikov_sinddm_2022}, which we include here for completeness.
Specifically, we compute CLIP text embeddings for each of the text prompts below (where ``\{\}'' denotes the prompt input), and compute the CLIP image embedding after applying each of the image augmentations below  to the input image. 
The CLIP loss is computed as the average cosine distance between these embeddings.
We use a patch size of three by three pixels in the diffusion process; empirically, we find that this patch size best captures textures and fine-scale image features, and the qualitative results align well with input text prompts.

\vspace{-1em}
\paragraph{Text augmentations.}
\begin{itemize}
\item ``photo of \{\}.'',
\item ``high quality photo of \{\}.'',
\item ``a photo of \{\}.'',
\item ``the photo of \{\}.'',
\item ``image of \{\}.'',
\item ``an image of \{\}.'',
\item ``high quality image of \{\}.'',
\item ``a high quality image of \{\}.'',
\item ``the \{\}.'',
\item ``a \{\}.'',
\item ``\{\}.'',
\item ``\{\}'',
\item ``\{\}!'',
\item ``\{\}...'',
\end{itemize}

\vspace{-1em}
\paragraph{Image augmentations.}
\begin{itemize}
\item Random spatial crops: 0.85 and 0.95 of the image size.
\item Random scaling: aspect ratio-preserving scaling, of both spatial dimensions.
\item Multiplication by a random factor, sampled uniformly from the range [0.8, 1.2].
\item Random horizontal-flipping is applied with probability p=0.5.
\item Random color-jittering: we jitter the global brightness, contrast, saturation and hue of the image.
\end{itemize}

\paragraph{Sampling.}
We use single-scale sampling for this task.  
We first run the inversion algorithm (Algorithm~\ref{alg:supp-inversion})
to map the clean input $\patch$ to an inverted noisy image
$\mathbf{x}_{t'}$ ($0 \rightarrow t'$).  
During the denoising phase ($t' \rightarrow 0$), 
we largely follow SinDDM~\cite{kulikov_sinddm_2022} for incorporating CLIP guidance, with some modifications.
At each timestep we compute the
denoised prediction $\denoisedpatch_t$ and apply the CLIP-guided update
\begin{equation}
    \denoisedpatch_{t,\text{CLIP}} \leftarrow
    \gamma \nabla_{\denoisedpatch_t}\mathcal{L}_{\text{CLIP}}
    + \lambda \denoisedpatch_t
    + (1 - \lambda)\denoisedpatch_{t+1,\text{CLIP}},
\end{equation}
where $\gamma$ controls CLIP guidance strength and $\lambda$ controls momentum
(i.e., retention of the previous estimate).
We use a time-dependent guidance schedule
$\gamma(t) = \frac{\sigma(t)}{\sigma(t')} \gamma(t') \in [0, \gamma(t')]$,
where $\gamma(t') = \gamma_{\text{max}}$ and the strength decays to $0$ as
$t$ approaches $0$.  
The momentum term helps prevent CLIP updates from being overridden by the
denoiser~\cite{kulikov_sinddm_2022}; we set $\lambda = 0.1$ for all
experiments. Also, following~\cite{kulikov_sinddm_2022}, we do not use the raw CLIP gradient but
instead compute
\begin{align}
    \mathbf{g}_t &\gets
        \nabla_{\lambda \denoisedpatch_t +
        (1 - \lambda)\denoisedpatch_{t+1,\text{CLIP}}}
        \mathcal{L}_{\text{CLIP}}, \\
    \mathbf{m}_t &\gets
        \mathbf{1}\bigl\{
            \text{pixel-norm}(\mathbf{g}_t)
            > (1 - f)\text{-quantile}
        \bigr\}, \\
    \mathbf{g}_t &\gets
        \|\denoisedpatch_t \odot \mathbf{m}_t\|_2
        \frac{
            \mathbf{g}_t \odot \mathbf{m}_t
        }{
            \|\mathbf{g}_t \odot \mathbf{m}_t\|_2
        },
\end{align}
where the \emph{fill factor} $f \in [0,1]$ specifies the fraction of pixels
allowed to be modified by the CLIP gradient at each step.  
The final normalization rescales the gradient so that its energy (within the
modified region) matches that of the denoised prediction.

\begin{figure*}[t]
    \centering
    \includegraphics[width=\textwidth]{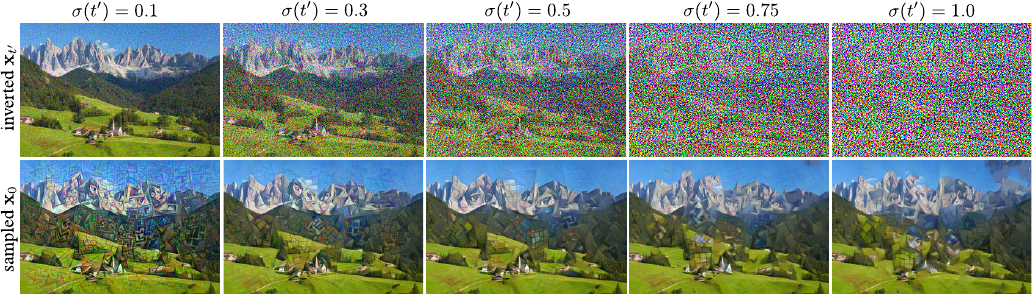}
    \caption{
    A sweep over inversion noise levels $\sigma(t') \in \{0.1, 0.3, 0.5, 0.75, 1.0\}$ using the
    prompt ``Cubism.''  
    The top row shows the inverted images $\noisypatch_{t'}$ obtained from the clean input $\mathbf{y}$ 
    ($0 \rightarrow t'$), and the bottom row shows the corresponding CLIP-guided
    denoised samples ($t' \rightarrow 0$).  
    We fix $\gamma(t') = \gamma_{\text{max}} = 0.5$ and $f = 1.0$, and we adjust the number of
    timesteps so that the change in noise level $\sigma(t) - \sigma(t{-}1)$ is constant;
    thus higher $\sigma(t')$ requires more inversion and denoising steps (e.g.,
    $\sigma(t') = 1.0$ corresponds to $t' = T = 200$ steps).  
    Across all settings, in particular for $\sigma(t') = 1.0$, the input structure is preserved due to the inversion algorithm.  
    Larger $\sigma(t')$ values (i.e., deeper inversion) reduce CLIP-induced noise in the final output,
    but also yield slightly weaker adherence to the original image.
    }
    \label{fig:supp-text-style-sweep-tprime}
\end{figure*}

\begin{figure*}[ht]
    \centering
    \includegraphics[width=\textwidth]{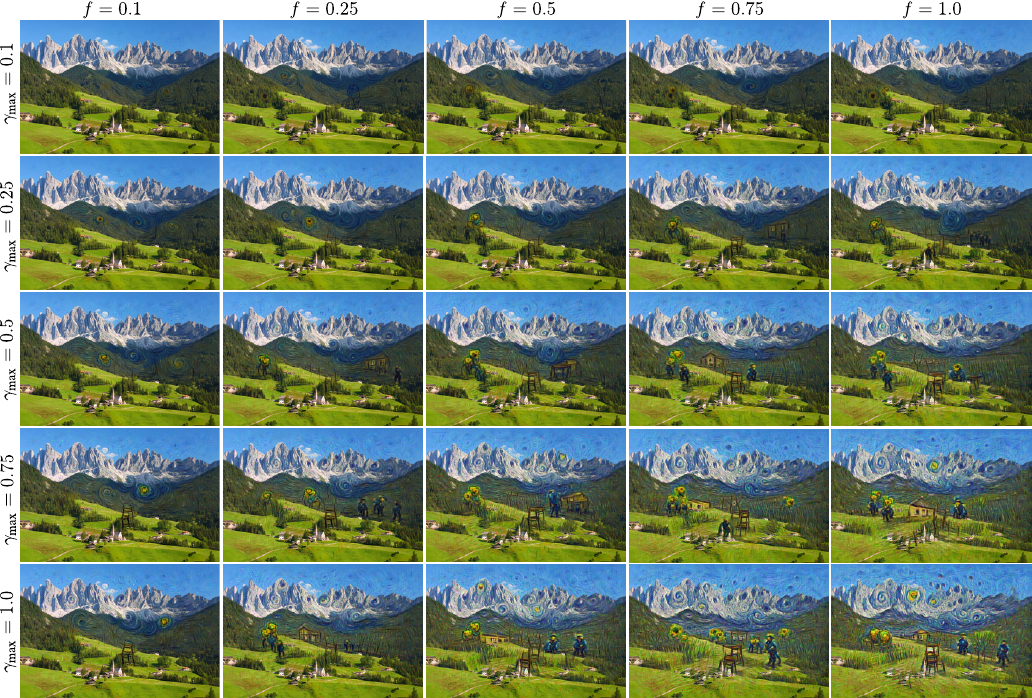}
    \caption{
    A sweep over guidance strength $\gamma_{\text{max}}$ and fill factor $f$.  
    All generations start from an inverted image at noise level $\sigma(t') = 0.3$
    and use 60 inversion and 60 denoising steps. 
    All generation uses prompt ``Van Gogh''.
    Stronger guidance yields more pronounced features aligned with the prompt,
    while larger fill factors cause edits to affect a greater portion of the image.
    }
    \label{fig:supp-text-style-sweep-gamma-vs-f}
\end{figure*}

We use a patch size of $7$ in all experiments.  
A sweep over $\sigma(t')$ is shown in
Figure~\ref{fig:supp-text-style-sweep-tprime}, demonstrating that style transfer is
achievable across all inversion depths.  
A sweep over $\gamma_{\text{max}}$ and $f$
(Figure~\ref{fig:supp-text-style-sweep-gamma-vs-f}) shows that larger values of both
parameters produce samples that more strongly follow the prompt.  
For comparison with SinDDM in Figure~\ref{fig:supp-text-style-transfer}, we set
$\gamma_{\text{max}} = 1.0$ and $f = 0.5$.

\clearpage
\begin{figure}[ht]
  \centering
  \includegraphics{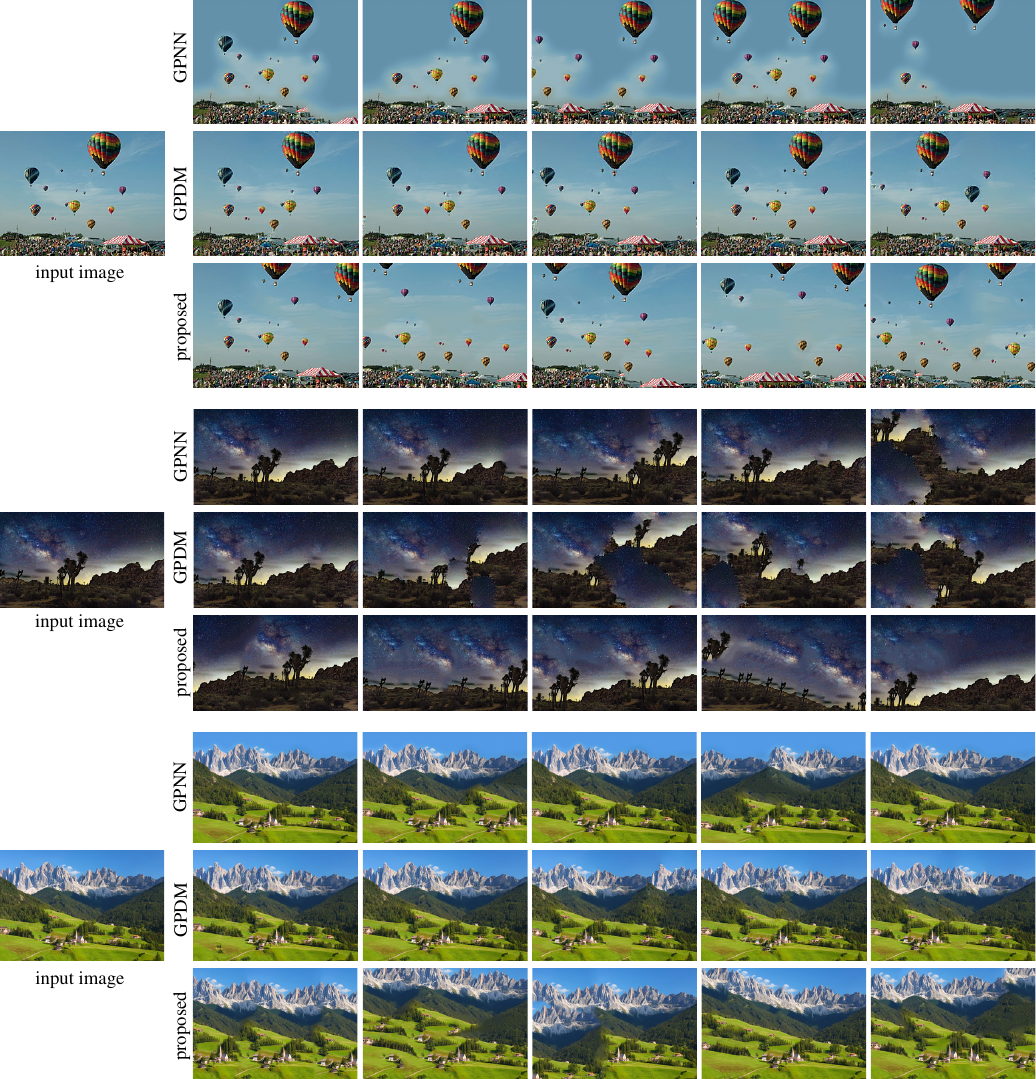}
  \vspace{-1.0em}
  \caption{Randomly sampled outputs from GPNN, GPDM, and our proposed method. GPNN and GPDM outputs are often nearly identical to the input (also see Table~\ref{tab:near_dups}).}
  \label{fig:duplicate_comparisons}
\end{figure}

\begin{table}[ht]
  \centering
  \captionsetup{justification=raggedright,singlelinecheck=false}
  \small
  \begin{tabular}{l|ccccccc}
  \toprule
  {Method}
  & {$r{=}100\%$}
  & {$r{=}95\%$}
  & {$r{=}90\%$}
  & {$r{=}85\%$}
  & {$r{=}80\%$}
  & {$r{=}75\%$}
  & {$r{=}70\%$}
  \\
  \midrule
  GPNN
  & 0.13\%
  & 10.27\%
  & 16.93\%
  & 25.60\%
  & 34.80\%
  & 43.60\%
  & 52.40\%
  \\
  GPDM 
  & 0.00\%
  & 8.27\%
  & 12.80\%
  & 21.20\%
  & 25.07\%
  & 32.67\%
  & 44.27\%
  \\
  proposed ($T=10$, $\eta = 0.0$)
  & 0.00\%
  & 0.00\%
  & 0.00\%
  & 0.13\%
  & 0.40\%
  & 0.67\%
  & 2.13\%
  \\
  \bottomrule
  \end{tabular}
  \caption{Near-duplicate rates at different $r\%$-pixel thresholds, 
  with average\ RGB difference ${<}15/255$ defining a near-duplicate.}
  \label{tab:near_dups}
  \vspace{-1em}
\end{table}

\begin{figure*}[ht]
    \centering
    \includegraphics[width=0.85\textwidth]{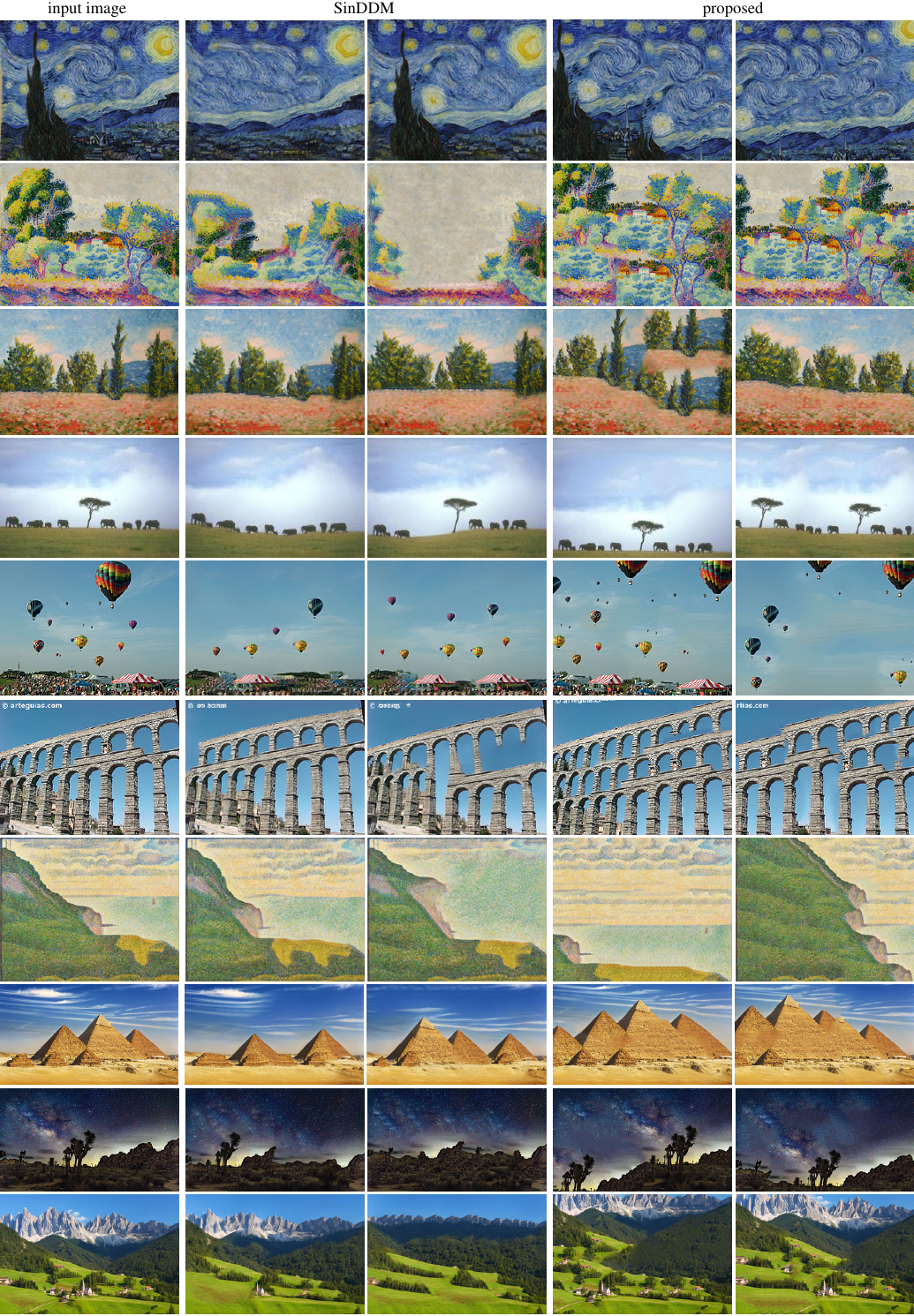}
    \caption{Additional comparisons to unconditional generation with SinDDM~\cite{kulikov_sinddm_2022} and the proposed method. Samples from the proposed method show similar quality and slightly greater diversity than the outputs of SinDDM, which agrees with the quantitative results shown in the main paper.}
    \label{fig:supp-unconditional}
\end{figure*}

\section{Supplementary Results}
\label{sec:supp-results}

\paragraph{Unconditional generation.}
The 15 images used for unconditional generation were downloaded from the SinDDM repository, with some originating from Places50 and SIGD16.
We compare against GPNN and GPDM in Figure~\ref{fig:duplicate_comparisons}.
While their samples often closely resemble the input image, our method produces more diverse outputs.
We quantify this using the near-duplicate ratio (Table~\ref{tab:near_dups}).
For each of the 15 input images, we sample 50 outputs for each method and report the fraction that closely match the input.
An output is considered a near-duplicate if at least $r\%$ of pixels have average RGB values within $15/255$ of the input.

Additional comparisons of unconditional generation with SinDDM~\cite{kulikov_sinddm_2022} and the proposed method are shown in Figure~\ref{fig:supp-unconditional}. Our approach demonstrates similar quality with appreciably more diversity than the SinDDM results, which agrees with the quantitative results shown in the main paper. 

\begin{figure*}[ht]
    \centering
    \includegraphics[width=1.0\textwidth]{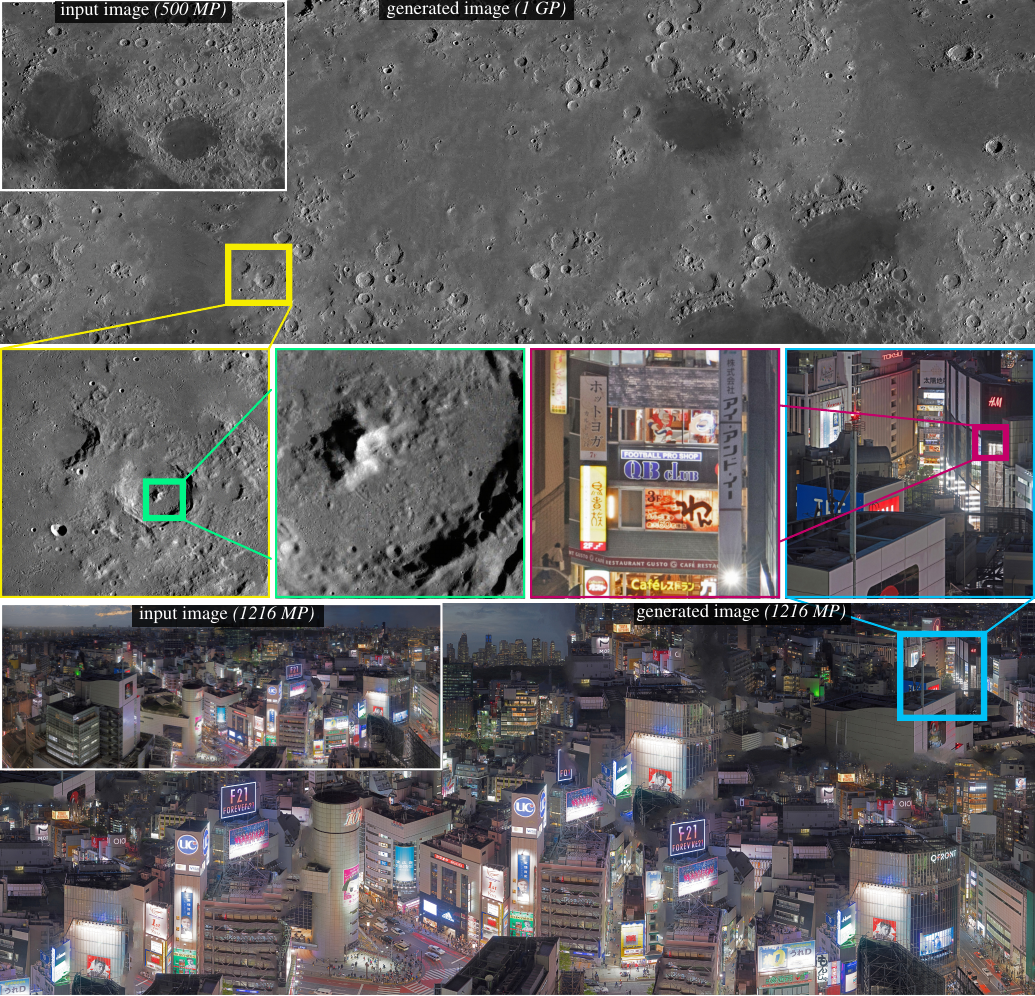}
    \caption{Supplemental gigapixel generation results. We use a combination of latent space diffusion, approximate nearest neighbors, and a fused attention kernel to generate these images in  33 minutes (top; \textit{Moon}) and 39 minutes (bottom; \textit{Tokyo}). \textit{Image credits: top — NASA (public domain); bottom — Trevor Dobson (CC BY–NC–ND).} }
    \label{fig:supp-high-res}
\end{figure*}

\paragraph{Gigapixel generation.}
We show additional examples of gigapixel image generation in Figure~\ref{fig:supp-high-res}. 
We generate these images in 33 minutes (\textit{Moon}) and 39 minutes (\textit{Tokyo}) on an NVIDIA RTX 6000 PRO by incorporating latent space diffusion, approximate nearest neighbors, and a fused attention operation.

\begin{figure}[ht]
    \centering
    \includegraphics[width=3.3in]{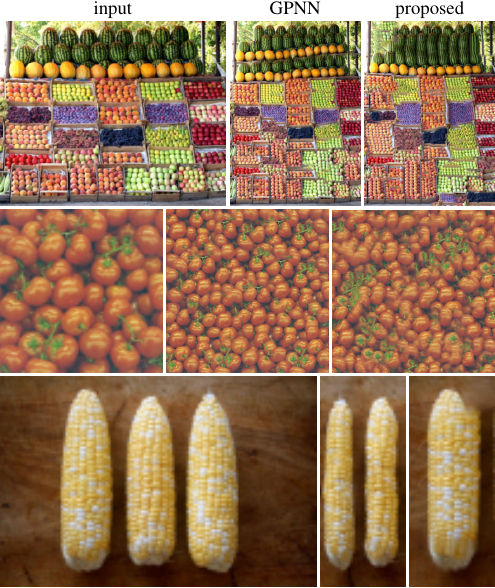}
    \caption{Image retargeting. We show the input image (leftmost column) the output of GPNN~\cite{granot_drop_2021} (middle column), and the results of the proposed method (right column). The proposed method  achieves results of similar quality to GPNN.}
    \label{fig:supp-retargeting-comparison}
    \vspace{-2em}
\end{figure}

\begin{figure*}[ht]
    \centering
    \includegraphics[width=0.85\textwidth]{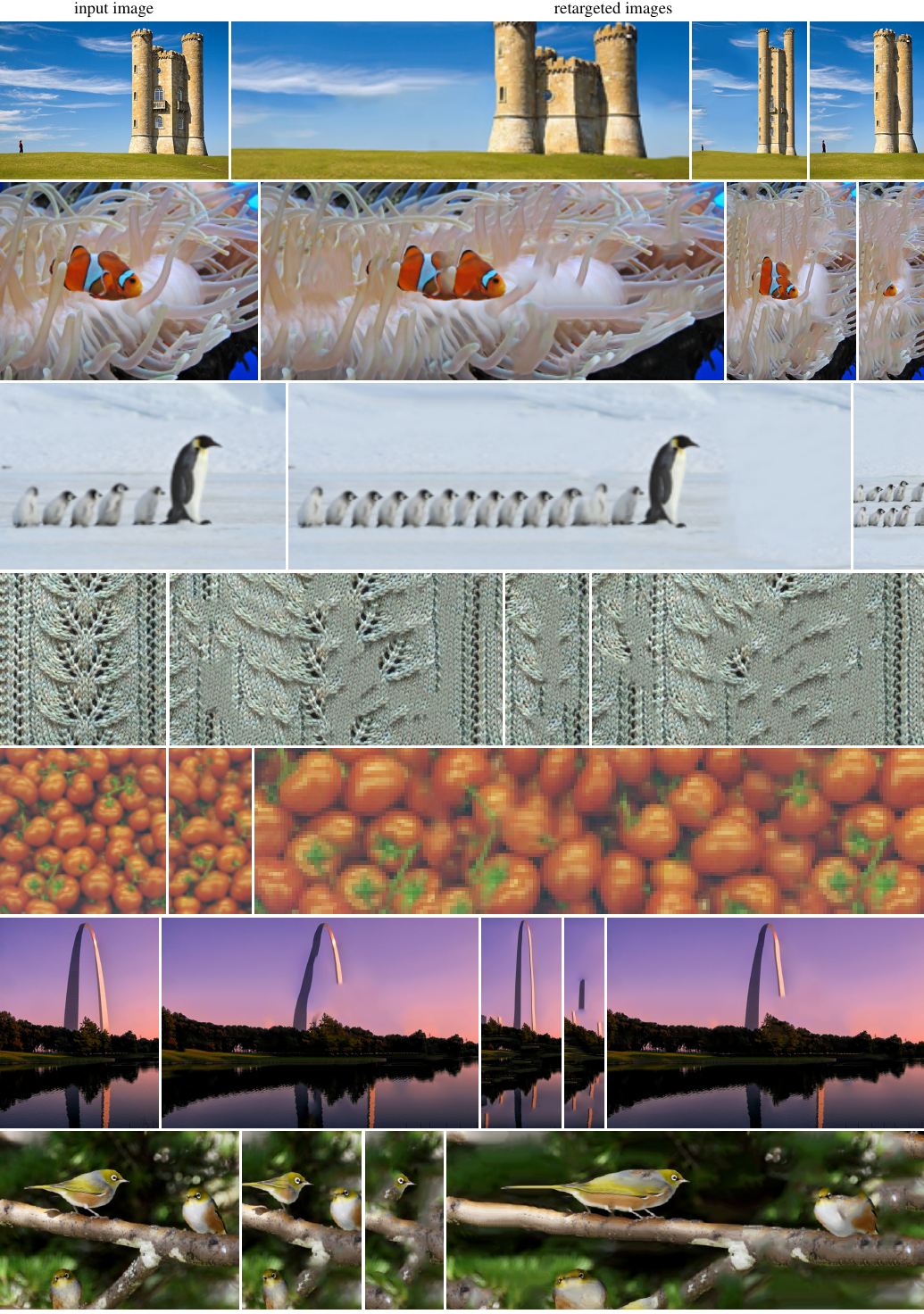}
    \caption{Supplementary results showing image retargeting with the proposed method. The input image is shown in the leftmost column of each row, and the other columns show retargeted samples.}
    \label{fig:supp-retargeting}
\end{figure*}

\paragraph{Image retargeting.}
We show a comparison between our method and GPNN in Figure~\ref{fig:supp-retargeting-comparison}.
Additional image retargeting examples are shown in Figure~\ref{fig:supp-retargeting}. 
Our method produces high-quality outputs across a variety of input images and aspect ratios.

\begin{figure*}[ht]
    \centering
    \includegraphics[width=0.9\textwidth]{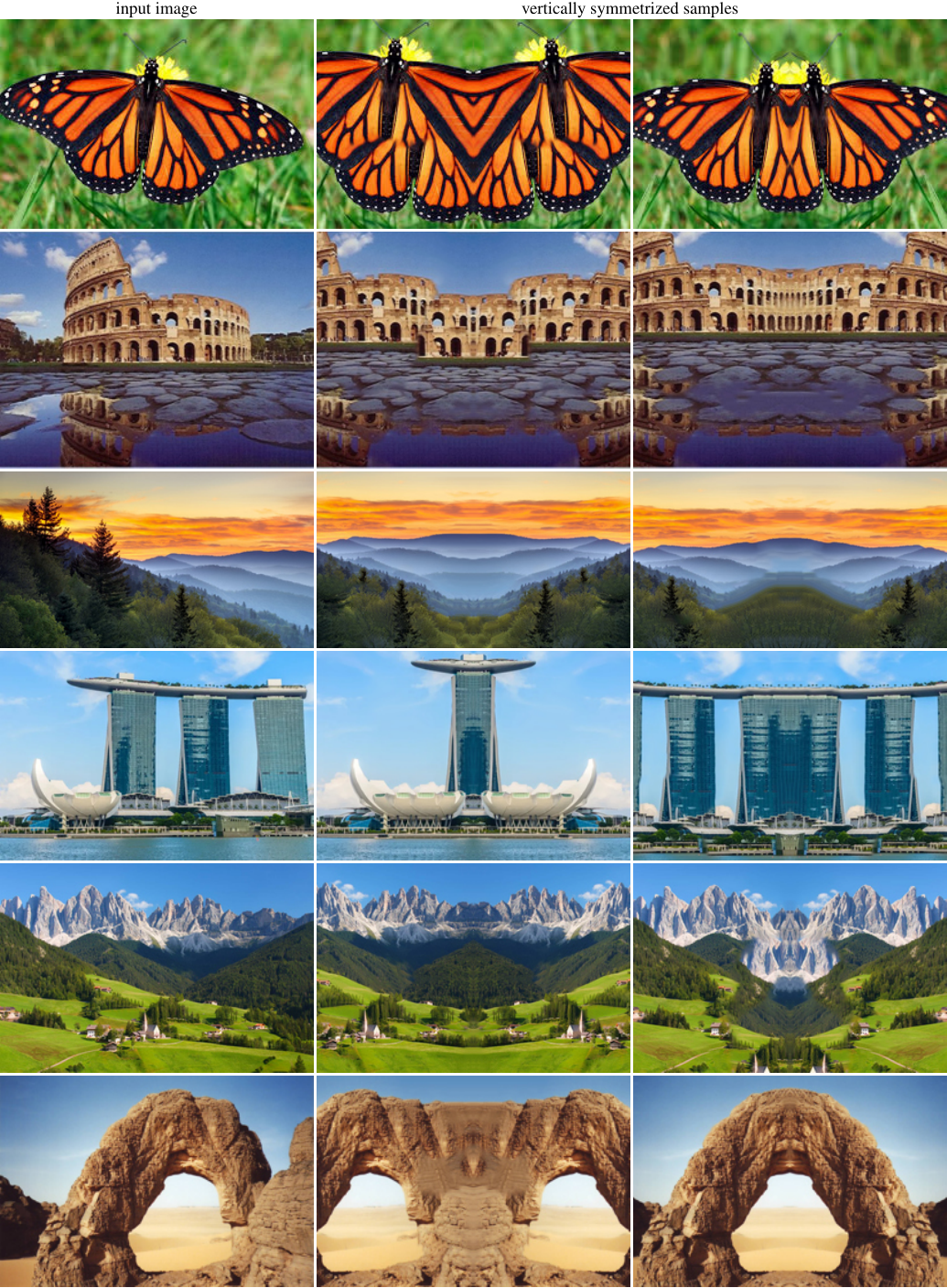}
    \caption{Supplementary results showing generated image samples with vertical symmetry.}
    \label{fig:supp-symmetrization}
\end{figure*}

\begin{figure*}[ht]
    \centering
    \includegraphics[width=0.9\textwidth]{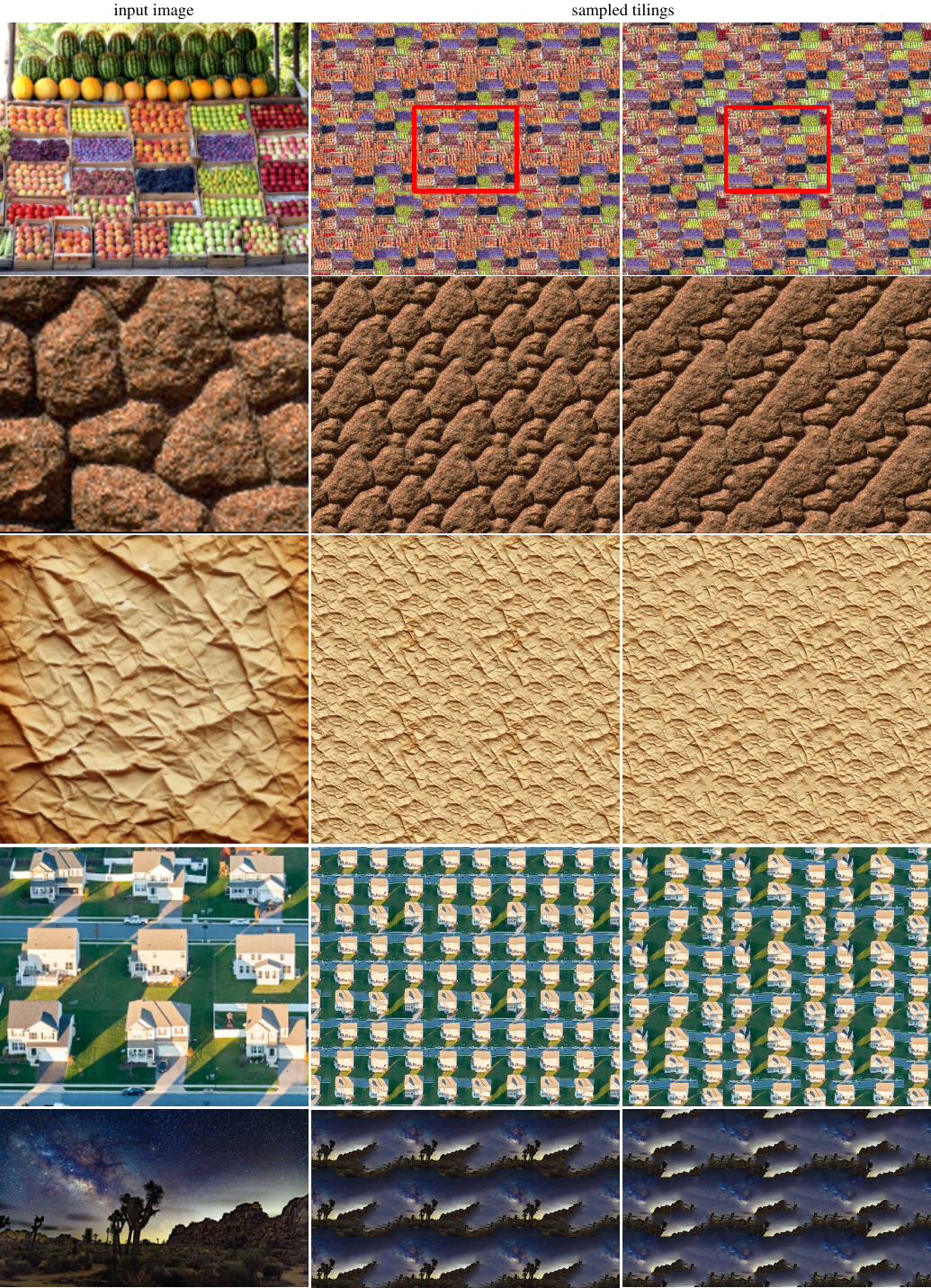}
    \caption{Supplementary results showing generated tileable image samples.}
    \label{fig:supp-tiling}
\end{figure*}

\vspace{-0.3em}
\paragraph{Image symmetrization.}
Additional generated samples with vertical image symmetry are shown in Figure~\ref{fig:supp-symmetrization}. We also provide additional generated tileable images in Figure~\ref{fig:supp-tiling}.

\begin{figure*}[ht]
    \centering
    \includegraphics[width=0.7\textwidth]{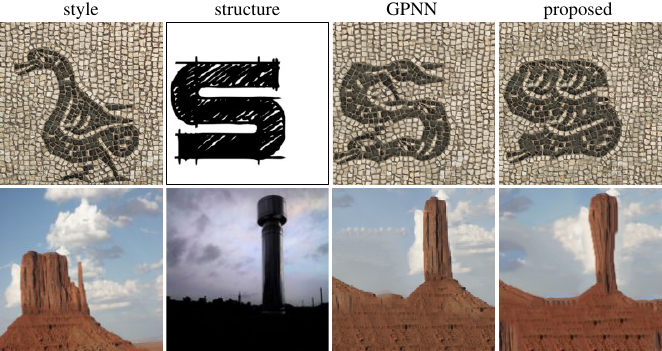}
    \caption{Supplementary results showing structural-analogies compared to GPNN. The proposed approach demonstrates qualitatively similar performance compared to GPNN.}
    \label{fig:supp-structural-analogies}
\end{figure*}

\vspace{-0.3em}
\paragraph{Structural analogies.}
Figure~\ref{fig:supp-structural-analogies} presents further structural analogy results alongside those produced by GPNN. 
The proposed approach offers similar qualitative performance to GPNN.

\begin{figure}[t]
  \centering
  \includegraphics{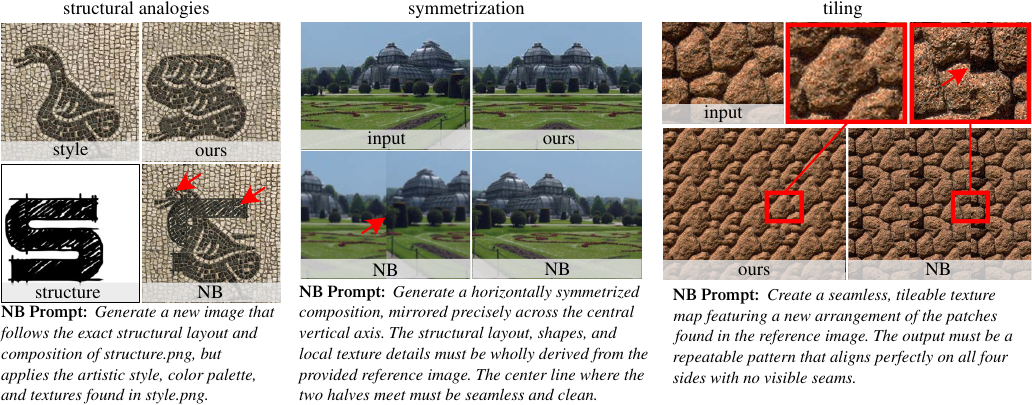}
  \caption{Nano Banana Pro (NB) results: outputs do not preserve the structure/input patch distribution (left), maintain symmetry (middle), or achieve seamless tiling (right).}
  \label{fig:large_model_comparision}
  \vspace{-1.0em}
\end{figure}

\vspace{-0.3em}
\paragraph{Comparison to large diffusion model.}
\label{sec:supp-comparisons-to-large-diffusion}
We compare our method against a large diffusion model, Nano Banana Pro~\cite{google2025nanobanana} (see Figure~\ref{fig:large_model_comparision}).
Because our approach generates output patches derived \emph{exclusively} from the input image, it provides a clear provenance chain between input and output—a guarantee that large models cannot offer.
Consequently, when Nano Banana Pro is applied to structural analogies, symmetrization, or tiling, its outputs fail to preserve the input patch distribution and do not satisfy the task constraints.

\begin{figure*}[ht]
    \centering
    \includegraphics[width=\textwidth]{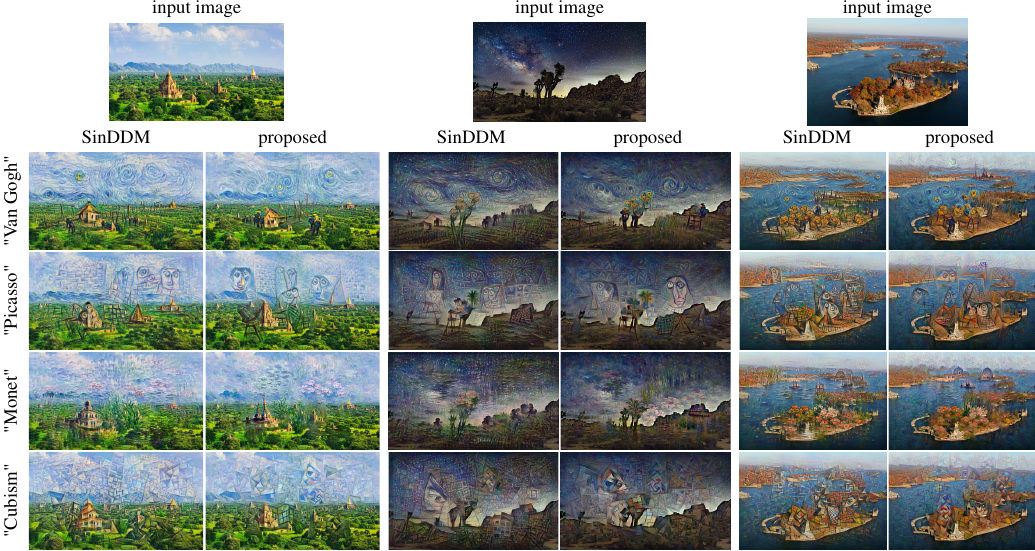}
    \caption{
    Supplementary results showing text-based style transfer compared to SinDDM. Patch size of $7 \times 7$, guidance of $\gamma_{\text{max}} = 1.0$, and fill factor $f=0.5$ are used for the proposed.
    The proposed approach demonstrates qualitatively similar performance compared to SinDDM for a range of prompts corresponding to different artists and styles.}
    \label{fig:supp-text-style-transfer}
\end{figure*}

\vspace{-0.3em}
\paragraph{Text-based style transfer.}
Additional text-based style transfer examples are provided in Figure~\ref{fig:supp-text-style-transfer}, together with outputs from SinDDM. 
Our method achieves comparable visual quality across a range of prompts spanning different artists and styles.

\begin{figure*}[ht]
    \centering
    \includegraphics[width=0.9\textwidth]{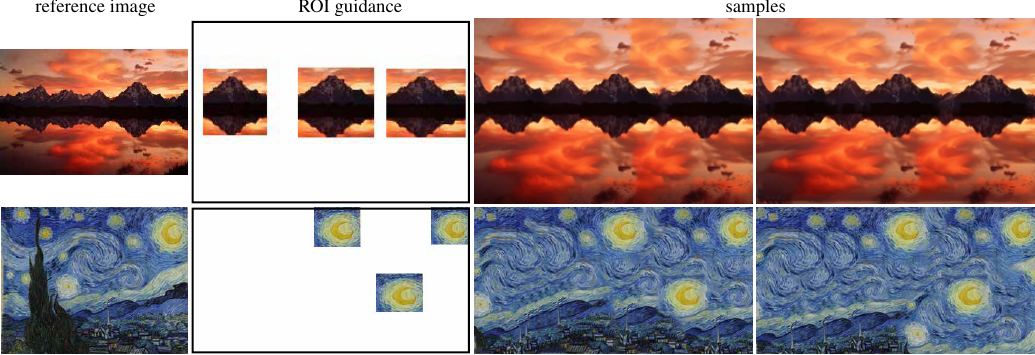}
    \caption{Supplementary results showing ROI-conditioned generation.}
    \label{fig:supp-roi}
\end{figure*}

\vspace{-0.3em}
\paragraph{Region of interest (ROI)-conditioned generation.}
After each denoising step we apply a mask to keep the provided ROI fixed, and retain the denoising prediction in the unmasked regions. 
We proceed in this fashion for all stages of the coarse-to-fine image sampling procedure except for the finest stage to avoid seams. 
Example results are shown in Figure~\ref{fig:supp-roi}.

\begin{figure*}[ht]
    \centering
    \includegraphics[width=0.9\textwidth]{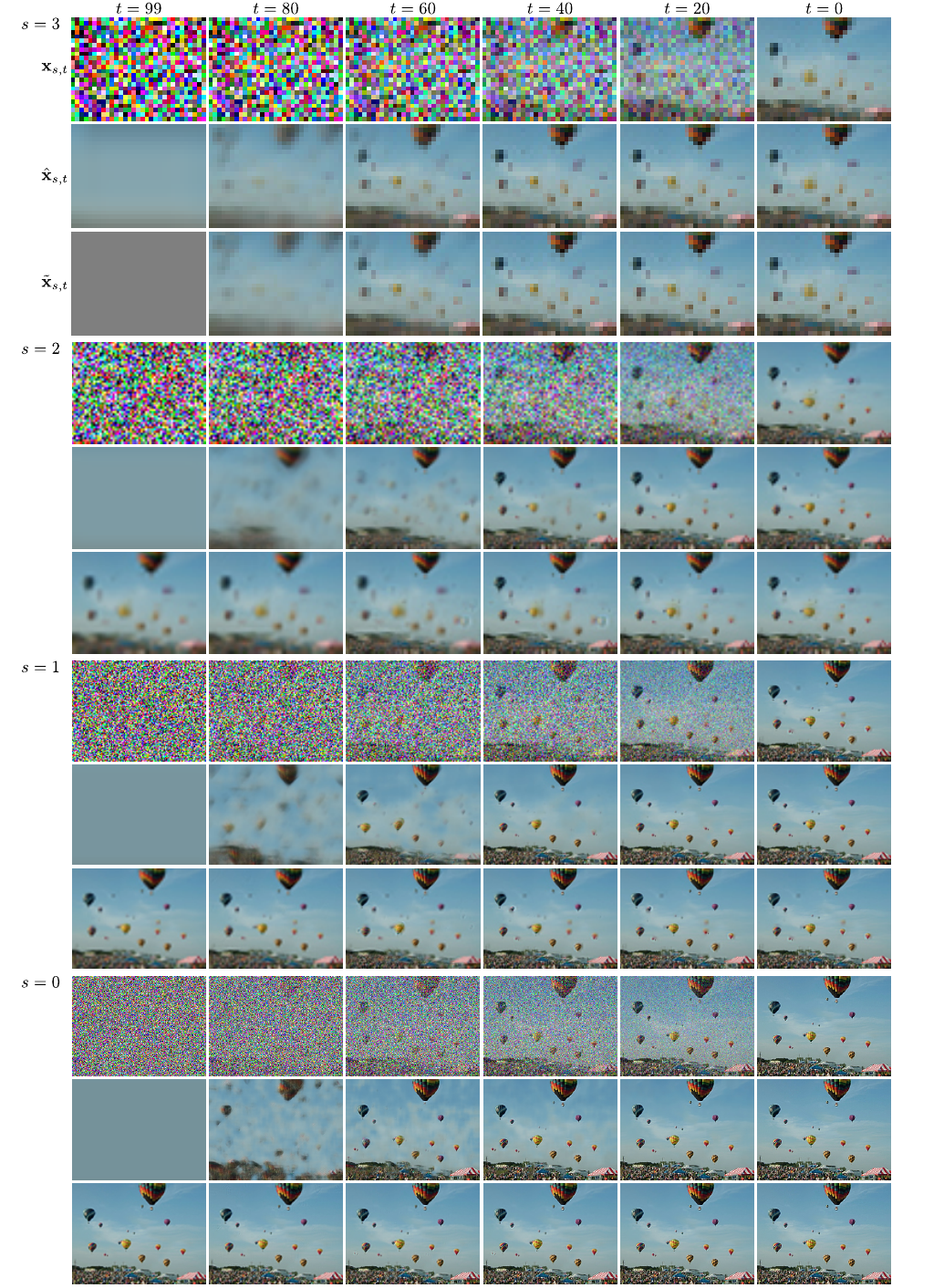}
    \caption{Visualization of coarse-to-fine image sampling for $S=4$ and $T=100$.}
    \label{fig:supp-gen-process}
\end{figure*}

\vspace{-0.3em}
\paragraph{Visualization of coarse-to-fine image sampling.}
We visualize the coarse-to-fine image sampling process in Figure~\ref{fig:supp-gen-process}.
The process is generated for $T=100$ and $S=4$, and we show images of $\noisypatch_{s, t}$, $\denoisedpatch_{s, t}$, and $\blendedpatch_{s, t}$ at $t=\{99, 80, 60, 40, 20, 0\}$ for all scales $s=\{3, 2, 1, 0\}$.  

\begin{figure}[t]
  \centering
  \includegraphics{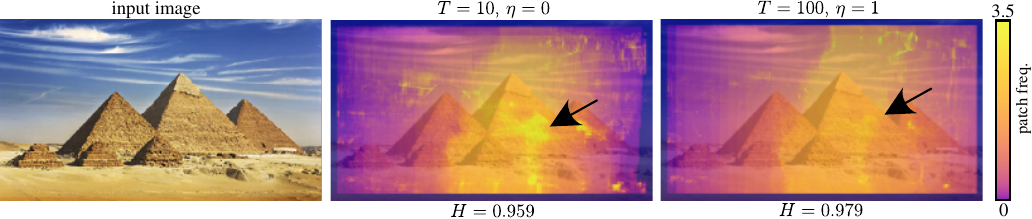}
  \caption{
    Patch frequency histogram \& entropy ($H$) for different $T$, $\eta$ for the unconditional generation. 
    Increasing $T$ and $\eta$ increases patch uniformity/entropy ($H$).
    }
  \label{fig:supp-entropy-analysis}
  \vspace{-1.0em}
\end{figure}

\paragraph{Patch sampling analysis.}
Given that at the final diffusion step ($\sigma(t) \to 0$), our method selects one patch from the dataset $\dataset$ for each patch location in the sampled image,
we can analyze the patch sampling distribution by counting the number of times each patch in $\dataset$ is selected across multiple samples.
Figure~\ref{fig:supp-entropy-analysis} shows histograms of the mean number of patch occurrences across 50 samples, overlaid on the input pyramid image (visualized at patch centers; borders are therefore zero).
The entropy value measures the uniformity of patch sampling, with higher values indicating more uniform sampling of patches (with a maximum of $1.0$, corresponding to perfectly uniform sampling).
Increasing $T$ and $\eta$ increases patch uniformity (entropy $H$), but also increases similarity to the input image.
For small $T$ and $\eta$, we hypothesize that image reconstruction from patches (Algorithm~\ref{alg:single-scale-sampling}, L14) constrains the layout at the coarsest scale, causing certain patches to be selected more frequently.

\clearpage

\clearpage
\section{Discussions and Extensions}
\label{sec:supp-discussion-and-extentions}

\subsection{Coarse-to-fine Sampling and Patch Size}
While our approach uses explicit multi-scale modeling, the diffusion process itself can be viewed as operating in a multi-scale fashion as noise is gradually added or removed~\cite{bansal2023cold, rissanen2023generative, hoogeboom2023blurring, dieleman2024spectral, mukhopadhyay2026scale}.
Hence, one open question is whether our explicit multi-scale approach can be modified, or even removed entirely and instead accommodated within the diffusion process.

    One straightforward modification of our approach is to exchange the scale--noise inner and outer loops in Algorithm~\ref{alg:coarse-to-fine}, i.e., first sample all scales at the first diffusion step, then move to the second diffusion step and sample all scales, and so on.
    This is largely equivalent to using multiple patch sizes to denoise at each noise level.
    We found that such unconditional generation samples have similar quality to those of our proposed algorithm, and intuitively, denoising steps at finer scales and higher noise levels (small $s$ and large $t$) contribute little to the final sample.
    
    Another alternative is to schedule the patch size during the diffusion process~\cite{niedoba2025mechanisticexplanationdiffusionmodel}.
    However, in the initial few steps, the patch size is on the order of the full image resolution, 
    which removes the efficiency benefits of using a constant small patch size across scales, as in our design.
    Alternatively, approaches such as the recent Scale-Space Diffusion model~\cite{mukhopadhyay2026scale}, which incorporates explicit downsampling schedules, could be integrated with our method.
We leave these directions for future work.

\subsection{Potential Extensions}
There are several interesting extensions to our method, which we briefly discuss below.

\paragraph{Multiple images as references.}
One can easily extend our method to take multiple images as input by simply concatenating the patch datasets of all input images into one patch dataset $\dataset$.
This enables generation that draws from multiple sources, e.g., blending textures across images or aggregating complementary views of the same scene.

\paragraph{Priors in data-scarce or out-of-distribution domains.}
Single-image and patch-based models could provide useful priors when datasets are limited or the target domain differs substantially from large-model training data (e.g., astronomy, hyperspectral imaging, radar). These priors could also aid inverse problems, e.g., via Diffusion Posterior Sampling~\cite{chung_diffusion_2023}.
In this setting, the image sample is updated by alternating between an image prior (in this case, our patch-based image denoiser could potentially be used instead of a trained large diffusion denoiser) and a likelihood step that constrains the sample to be consistent with the observed measurements.

\paragraph{Complementarity with large diffusion priors.}

We envision that one could combine latent multi-image patch-based models with large-scale latent diffusion models by incorporating patch-based image priors as an additional guidance signal alongside standard text conditioning. 
This formulation would leverage the complementary strengths of large diffusion models—such as semantic understanding and global coherence—and patch-based models, which capture fine-scale image statistics from the input images. 
By adjusting the relative influence of these guidance signals, the generation could be steered to reflect both global semantics and local image structure. We see this as a particularly exciting future direction.

\end{document}